\documentclass{article}

\usepackage{PRIMEarxiv}
\def\keywordname{{\bfseries \emph{Keywords}}}%
\def\keywords#1{\par\addvspace\medskipamount{\rightskip=0pt plus1cm
\def\and{\ifhmode\unskip\nobreak\fi\ $\cdot$
}\noindent\keywordname\enspace\ignorespaces#1\par}}

\usepackage[numbers,sort]{natbib}
\usepackage[utf8]{inputenc} 
\usepackage[T1]{fontenc}    
\usepackage{hyperref}       
\usepackage{url}            
\usepackage{booktabs}       
\usepackage{amsfonts}       
\usepackage{nicefrac}       
\usepackage{microtype}      
\usepackage{lipsum}
\usepackage{fancyhdr}       
\usepackage{graphicx}       
\graphicspath{{media/}}     

\usepackage[skip=1pt]{caption}

\usepackage{multirow}
\usepackage{array}
\usepackage{bbm}
\usepackage[skip=2pt]{caption}
\usepackage{subcaption}
\usepackage[inline]{enumitem}
\usepackage{algorithmic}
\usepackage{graphicx}

\usepackage{comment}
\usepackage[linesnumbered,ruled,vlined]{algorithm2e}
\usepackage{physics}
\usepackage{color} 

\SetKwComment{Comment}{$\triangleright$\ }{}
\SetKwRepeat{Do}{do}{while}

\newcolumntype{C}[1]{>{\centering\let\newline\\\arraybackslash\hspace{0pt}}m{#1}}
\newcommand{\ours}{\textbf{LTG}}

\title{Learning Latent Spaces for Domain Generalization\\ in Time Series Forecasting
}

\author{
  Songgaojun Deng \\
  University of Amsterdam \\
  Amsterdam, The Netherlands\\
  \texttt{s.deng@uva.nl} \\
   \And
  Maarten de Rijke \\
  University of Amsterdam \\
  Amsterdam, The Netherlands\\
  \texttt{m.derijke@uva.nl} \\
}

\begin{document}
\maketitle

\begin{abstract}
Time series forecasting is vital in many real-world applications, yet developing models that generalize well on unseen relevant domains -- such as forecasting web traffic data on new platforms/websites or estimating e-commerce demand in new regions -- remains underexplored. 
Existing forecasting models often struggle with domain shifts in time series data, as the temporal patterns involve complex components like trends, seasonality, etc. While some prior work addresses this by matching feature distributions across domains or disentangling domain-shared features using label information, they fail to reveal insights into the latent temporal dependencies, which are critical for identifying common patterns across domains and achieving generalization. 

We propose a framework for domain generalization in time series forecasting by mining the latent factors that govern temporal dependencies across domains. 
Our approach uses a decomposition-based architecture with a new Conditional $\beta$-Variational Autoencoder (VAE), wherein time series data is first decomposed into trend-cyclical and seasonal components, each modeled independently through separate $\beta$-VAE modules. The $\beta$-VAE aims to capture disentangled latent factors that control temporal dependencies across domains. We enhance the learning of domain-specific information with a decoder-conditional design and introduce domain regularization to improve the separation of domain-shared and domain-specific latent factors.
Our proposed method is flexible and can be applied to various time series forecasting models, enabling effective domain generalization with simplicity and efficiency. 
We validate its effectiveness on five real-world time series datasets, covering web traffic, e-commerce, finance and power consumption, demonstrating improved generalization performance over state-of-the-art methods.
\end{abstract}

{\keywords{Time series forecasting, Domain generalization}}

\maketitle

\section{Introduction}
Time series forecasting is essential for various online applications, including web traffic analysis, e-commerce, finance, and digital healthcare services.
It focuses on predicting future movements based on past observations, making accurate forecasting vital for informed decision-making and resource allocation. Effective forecasting can lead to significant benefits, such as optimized websites, improved supply chain management, enhanced user experiences, and better public health responses.

A major challenge in time series forecasting is data distribution shifts~\citep{gagnon2022woods}, including non-stationarity and domain shifts. In the context of the web, these shifts are exacerbated by rapidly changing online environments and user behaviors. While much research addresses non-stationarity~\citep{kim2021reversible,fan2023dish,liu2024adaptive}, domain shifts -- arising from variations in data sources, platforms, or user interactions -- are less explored~\cite{deng2024domain}.
This challenge is crucial in applications such as allocating server resources for new websites, forecasting demand for new products, or analyzing user behavior across demographics.

Domain generalization aims to develop models that can generalize to unseen domains based on training on a set of relevant source domains. While substantial advances have been made in computer vision~\cite{yue2019domain,gong2019dlow} and natural language processing~\cite{balaji2018metareg,hupkes2023taxonomy}, generalizing across time series domains remains challenging due to factors like higher uncertainty and dynamic temporal characteristics~\cite{du2021adarnn}.
In time series forecasting, existing studies (i)~primarily use feature alignment/matching~\cite{du2021adarnn,zhang2022domain,deng2024domain} to promote common feature learning for domain generalization across different sources (i.e., distribution shifts over data sources) or (ii)~generate time-sensitive networks~\cite{bai2023temporal,nasery2021training} to address generalization over time domains (i.e., distribution shifts over time)~\cite{gagnon2022woods}. 
While it is widely recognized that learning temporal dependencies is crucial for effective sequential modeling~\cite{hochreiter1997long,chung2014empirical,vaswani2017attention}, prior approaches often neglect the latent factors that control temporal dependencies across domains -- whether shared or domain-specific. 
We are motivated to address this gap by taking one step to explore the underlying dynamics in temporal dependencies to enhance generalization.

Modeling latent dynamics in time series can help identify common patterns shared across domains, improving generalization to unseen domains with similar temporal characteristics.
Nevertheless, it presents several challenges: (i)~Explicitly capturing relationships across timestamps is difficult due to the dynamic nature and inherent uncertainty of time series data. Meanwhile, the common windowing process used in time series analysis often leads to information loss (e.g., missing global information), making it challenging to capture temporal dependencies statically.
(ii)~Temporal dependencies are not easily quantifiable, especially considering the various components (trends, seasonality) in time series data and the complexities involved in designing recurrent units~\cite{chung2014empirical}. 
(iii) Although some latent information can be implicitly captured -- such as through variational autoencoders (VAEs)~\cite{kingma2013auto,higgins2017beta,zhao2023revisiting,wang2024revisiting}, which excel at capturing latent information -- effectively capturing both domain-specific and domain-shared information to improve generalization across unseen domains remains a challenge.

To address the challenges mentioned above, we introduce a framework, \textbf{Latent Temporal Generalization (LTG)}, which models the underlying temporal dependencies in latent spaces to achieve domain generalization in time series forecasting. 
The core idea is to learn latent vectors sampled from a latent space that comprises both domain-shared and domain-specific components. They imply temporal dependencies within and across domains and support generalization to unseen domains in forecasting tasks. 
Our method employs a new Conditional $\beta$-VAE to estimate the latent factors both within and across domains for each time series, guiding a forecasting decoder to predict future steps in unseen domains. 
To enhance latent factor learning in complex time series data, we first decompose the time series into trend-cyclical and seasonal components, allowing the $\beta$-VAE to reconstruct each part of the input sequence independently. 
The $\beta$-VAE decoder is conditioned on domain identifiers to capture domain-specific information while keeping the encoder domain-agnostic.
In addition, we introduce domain regularization to encourage the separation of domain-shared and domain-specific latent factors.

This work makes the following key contributions:
\begin{itemize}
    \item We introduce a framework, \ours{}, for domain generalization in time series forecasting, which models latent factors that capture temporal dependencies both within and across domains.
    \item We present an effective Conditional $\beta$-VAE, which conditions the decoder on domain identifiers to capture domain-specific information in the latent space. 
    \item We enhance the latent factors with a domain regularization technique that better regulates the separation between domain-shared and domain-specific latent factors.
\end{itemize}
We report on extensive experiments on five real-world datasets across diverse application domains, and the results indicate that our method surpasses existing state-of-the-art approaches in domain generalization tasks.

\section{Related Work}
\subsection{Time Series Forecasting}
Time series forecasting has been extensively studied in numerous real-world applications.  Classical approaches like autoregressive (AR)~\cite{tibshirani1996regression}, ARIMA~\cite{benjamin2003generalized}, and exponential smoothing~\cite{gardner2006exponential}, focus on modeling linear patterns in time series data. To capture non-linear relationships and more complex temporal dependencies, machine learning approaches such as regression~\cite{tibshirani1996regression} and tree-based models~\cite{ke2017lightgbm} have been explored. In recent years, deep .learning methods have gained significant attention for their ability to capture these complexities. 
These include recurrent neural networks (RNNs) and their variants~\cite{hochreiter1997long,chung2014empirical}, as well as temporal convolutional networks~\cite{bai2018empirical,oord2016wavenet}, which offer parallelization benefits over RNNs. 
Moreover, attention-based models~\cite{bahdanau2014neural,vaswani2017attention,li2019enhancing}, such as Informer~\cite{zhou2021informer}, FEDformer~\cite{zhou2022fedformer}, and PatchTST~\cite{nie2023a}, have achieved state-of-the-art performance in many time series forecasting tasks. 
The success of large language models (LLMs)~\cite{radford2019language,zhao2023survey} has inspired researchers to use pre-trained LLMs for time series forecasting, resulting in advanced models like GPT4TS~\cite{zhou2023one}, TimeLLM~\cite{jin2024timellm} and UniTime~\cite{liu2024unitime}. New insights are emerging regarding the actual effectiveness of LLMs in time series tasks, particularly considering their inherent language knowledge misaligned with time series data and high computational demands~\cite{tan2024language}.
Additionally, large-scale time series models~\cite{ansari2024chronos,woo2024unified}, with LLM backbones trained on vast time series datasets, have shown strong performance across diverse time series tasks. 
Forecasting methods can be classified into point forecasting and probabilistic forecasting~\cite{gneiting2014probabilistic,lim2021time}, with the latter being critical for decision-making due to their ability to quantify uncertainties~\cite{gneiting2014probabilistic}. In this work, we focus on probabilistic forecasting and aim to improve the generalization of time series models across domains.



\subsection{Out-of-distribution Generalization in Time Series} 
Deep learning models often fail to generalize well under distribution shifts, driving research into out-of-distribution (OOD) generalization. 
OOD generalization has been studied extensively in static computer vision tasks~\cite{liu2021towards,wang2022generalizingunseendomainssurvey}. Recently, its application in time series has gained increasing attention due to common data distribution shifts in real-world scenarios~\cite{gagnon2022woods}. 
OOD generalization in time series can be categorized into domain generalization and subpopulation shift~\cite{gagnon2022woods}. Among these, domain generalization has garnered more attention and is further divided into generalization across \textit{time} domains or \textit{source} domains, depending on whether the shifts occur over time or across data sources.
In time-domain generalization, researchers have investigated time-sensitive networks that use parameter generation~\cite{bai2023temporal} to adapt to dynamic temporal behaviors in new domains, as well as gradient-based methods~\cite{nasery2021training} to facilitate smooth transitions in model behavior over time.
Other approaches involve dynamically identifying distribution domains and learning common patterns across them using distribution matching~\cite{du2021adarnn} or adversarial learning~\cite{lu2023outofdistribution}.
In source domain generalization, most studies focus on classification tasks, using label information to achieve domain invariance through methods like distribution matching~\cite{zhang2022domain}, data augmentation~\cite{zhang2022domain}, and contrastive learning~\cite{hu2022causal,ragab2022conditional}.
There is a recent study on domain generalization in time series forecasting, which uses cross-domain regularization to enhance the learning of common patterns via temporal feature distribution matching~\cite{deng2024domain}. Our study aligns closely with this research but aims to delve deeper into the latent factors that drive temporal dependencies for improved generalization. 
We want to highlight the distinction between our work and the recent study, UniTime~\cite{liu2024unitime}. While UniTime focuses on cross-domain learning, our approach emphasizes generalization to unseen domains.

\section{Problem Formulation}
This study focuses on domain generalization in time series forecasting~\cite{gagnon2022woods,deng2024domain}. We formulate the problem beginning with a general overview of time series forecasting.

\noindent\textbf{Time series forecasting.} 
We define a univariate time series as $\mathbf{y}_{1:T}=\{ y_1, y_2, \ldots, y_T\}$, where $y_t$ is the observed value at time $t$ (e.g., website traffic), and $T$ is the length of the historical/lookback window. The time step $t$ is usually constant (e.g., measured in hours or days). The goal of time series forecasting is to predict future values $\mathbf{y}_{T+1:T+h} = \{{y}_{T+1}, {y}_{T+2}, \ldots, {y}_{T+h} \}$, where $h\geq 1$ denotes the forecasting horizon.\footnote{We focus on univariant analysis and leave empirical verification and extension on multivariate settings for future studies.}
A time series dataset $D = \{ X, Y\}$ consisting of $N$ samples can be expressed as
$D = \{(\mathbf{y}_{i, 1:T}, \mathbf{a}_{i, 1:T}), \mathbf{y}_{i, T+1:T+h}  \}^N_{i=1}$,
where $\mathbf{a}$ denotes any external factors (e.g., categorical features). For simplicity, we use $\mathbf{x},\mathbf{a}$ to indicate input features and $\mathbf{y}$ for output time series in the following discussion. In this work, we focus on probabilistic forecasting~\cite{salinas2020deepar}, estimating the probability distribution of the time series' future, which captures uncertainties and provides more robust insights for real-world data.

\noindent\textbf{Domain generalization in time series forecasting.}
Let $\mathcal{D} = \{D^{1}, D^2, \ldots, D^K\}$ represent the set of $K$ domains, where each domain $D^j$ consists of a time series dataset $D^{j} = \{X^{j}, Y^{j}\}$.
We have access to $M$ training domains $\mathcal{D}_{\text{train}} = \{D^j\}_{j=1}^M$ where $M < K$. The goal is to learn a time series forecaster that performs well to unseen test domains $\mathcal{D}_{\text{test}} =\{D^{j}\}_{j=M+1}^{K}$. 

We consider the \textit{common underlying patterns} assumption~\cite{deng2024domain}, which posits that the existence of shared patterns across domains supports the relevance of generalization tasks. This is natural and we experiment with domains pertinent to specific fields, such as the various country domains of a website.\footnote{We do not rely on the assumption of \textit{no abrupt distribution shifts} from prior work~\cite{deng2024domain} and it is also difficult to quantify such shifts in real-world data. We also alleviate this effect using a reversible instance norm~\cite{kim2021reversible}.}

\begin{figure*}[t]
    \centering
    \includegraphics[width=\linewidth]{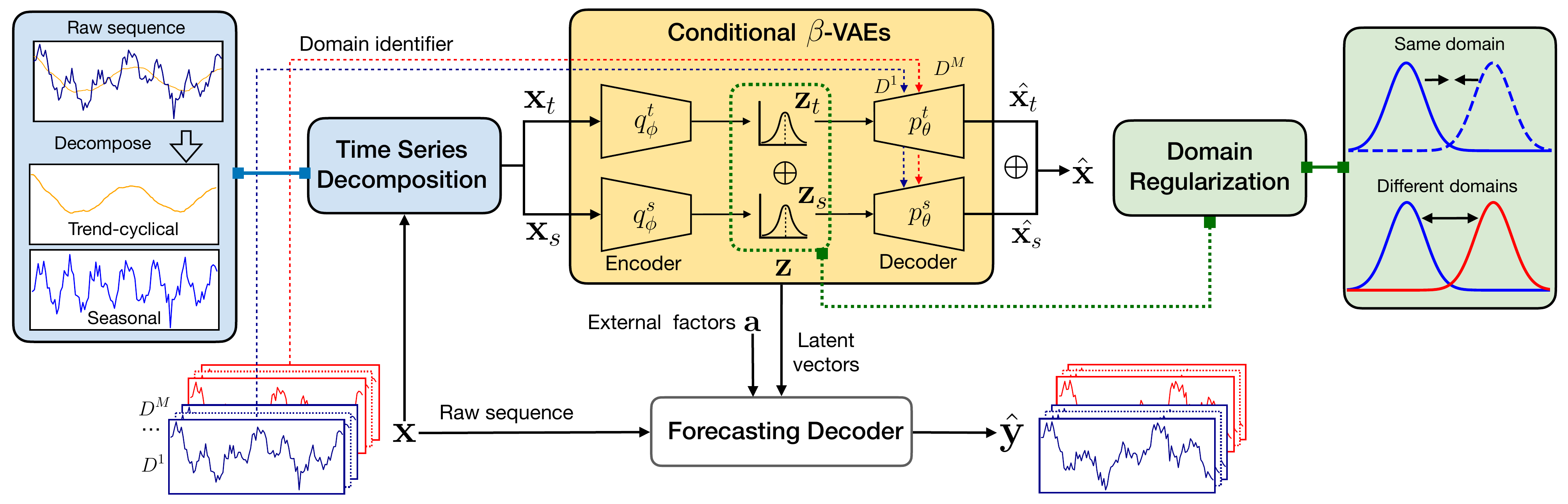}
    \caption{The proposed framework, \ours, consists of three components: (1) Time Series Decomposition, which decomposes a raw time sequence into trend-cyclical and seasonal components for more effective modeling of distinct temporal patterns. (2) Latent Factor Learning with Conditional $\beta$-VAE, which learns latent representations that capture temporal dependencies across different domains and ideally achieves disentangled latent factors for each dimension. (3) Domain Regularization, which regularizes domain-specific and shared latent features to enhance the disentangling of latent factors. The forecasting decoder is a flexible model that integrates the raw sequence, any external factors, and the latent vectors learned by the Conditional $\beta$-VAE to generate time series forecasts.
    }
    \label{fig:framework}
\end{figure*}

\section{Methodology}
We propose a novel framework for domain generalization in time series forecasting, \ours{}, illustrated in Figure~\ref{fig:framework}. The framework comprises three key components: 
(1) \textbf{Time Series Decomposition}, which breaks down a complex time series into its trend-cyclical and seasonal components. (2) Latent Factor Learning with \textbf{Conditional $\beta$-VAE}. This step learns latent factors from the decomposed time series components through a conditional $\beta$-VAE, where the decoder is conditioned on domain identifiers. This helps model the (disentangled) latent factors that capture temporal dependencies across different domains. (3) \textbf{Domain Regularization}, which further regularizes and separates domain-shared and specific latent factors.
Our framework integrates with a forecasting decoder, which uses both the input data and the learned latent representations to generate predictions and it can be any time series forecaster.
Below, we illustrate each component of our approach in detail.


\subsection{Time Series Decomposition}
Time series data often exhibit multiple components, such as trend, seasonality, and noise~\cite{hamilton2020time}. 
To effectively analyze and predict such data, normalization (e.g., zero-mean) is commonly used, and decomposition techniques have proven effective~\cite{cleveland1990stl,hamilton2020time}.
Recently, \citet{wu2021autoformer}~demonstrated that incorporating seasonal-trend decomposition progressively into forecasting frameworks enhances the predictability of raw data. Further advances include complex methods that combine trend components with various moving average kernels~\cite{zhou2022fedformer}.
In this work, we adopt a simple approach by performing decomposition at the initial stage~\cite{zeng2023transformers}. 
Specifically, we decompose the raw time series $\mathbf{x}$ into two components: trend-cyclical $\mathbf{x}_t$ and seasonal $\mathbf{x}_s$ components.\footnote{Note that we do not involve external features $\mathbf{a}$ in this process.} The trend-cyclical component $\mathbf{x}_t$ captures the long-term progression, while the seasonal component $\mathbf{x}_s$ reflects periodic patterns. The procedure is as follows:
\begin{align}
\mathbf{x}_t &= \text{AvgPool} \big (\text{Padding}(\mathbf{x_t}) \big ), \\
\mathbf{x}_s &= \mathbf{x} - \mathbf{x}_t,
\end{align}
where the AvgPool(·) computes the moving average while applying padding to maintain the original length of the series. The primary benefit of time series decomposition is that it simplifies the process of learning latent factors in subsequent steps, particularly when dealing with complex time series data. These latent factors are expected to capture essential information about underlying temporal dependencies, thereby facilitating generalization across domains.

\subsection{Latent Factor Learning with Conditional $\beta$-VAE}

To achieve domain generalization, capturing the common patterns across domains is essential. Some prior work attempts this by aligning feature distributions in time series data, aiming to ensure consistent patterns between input and output sequences in different domains~\cite{du2021adarnn,zhang2022domain,deng2024domain}. We argue that merely aligning learned temporal features is insufficient to capture the common patterns, as these methods provide limited insights into the underlying temporal dependencies.

To better generalize across domains, we aim to understand how past time steps influence future steps and identify where the common patterns lie across domains. A straightforward approach is to explicitly model these mechanisms by generating parameters for temporal models~\cite{bai2023temporal}, such as recurrent units, convolutional layers, and self-attention mechanisms. However, this approach can lead to over-parameterization, resulting in suboptimal performance. Instead, we propose to model temporal dependencies implicitly by learning latent factors that capture these relationships.

To this end, we introduce a novel Conditional $\beta$-VAE that learns latent temporal dependencies. It has a special decoder conditioned on domain identifiers, allowing the latent vectors to capture domain-specific information while maintaining the encoder domain-agnostic and estimating more general information. 
Following the time series decomposition, we model each component (trend-cyclical and seasonal) independently using a Conditional $\beta$-VAE  to capture latent factors regarding trend-cyclical and seasonal patterns $\mathbf{z}_t, \mathbf{z}_s$, respectively.
We present the proposed Conditional $\beta$-VAE below, based on an encoder-decoder structure.

\paragraph{\textbf{Encoder}} Two encoders learn to approximate the posterior distribution $q_{\phi_t}(\mathbf{z}_t|\mathbf{x}_t), q_{\phi_s}(\mathbf{z}_s|\mathbf{x}_s)$ of the latent variables $\mathbf{z}_t, \mathbf{z}_s$ given the decomposed trend-cyclical and seasonal components $\mathbf{x}_t, \mathbf{x}_s$.
\begin{equation}
    \mathbf{z}_t \sim   q_{\phi_t}(\mathbf{z}_t|\mathbf{x}_t), \qquad \mathbf{z}_s \sim   q_{\phi_s}(\mathbf{z}_s|\mathbf{x}_s).
\end{equation}
%
The latent variables $\mathbf{z}_t, \mathbf{z}_s \in R^{d_z}$ are generated from some prior distributions of the underlying temporal dependencies, typically chosen to be a Gaussian distribution:
\begin{equation}
    p(\mathbf{z}_t) = \mathcal{N}(0,\textbf{\textit{I}}), \qquad     p(\mathbf{z}_s) = \mathcal{N}(0,\textbf{\textit{I}}),
\end{equation}
where $\textbf{\textit{I}}$ is the identity matrix; $d_z$ is the dimension of the latent representation.  

\paragraph{\textbf{Decoder}} 
Two decoders reconstruct the decomposed trend-cycli\-cal and seasonal components using the latent vector and the domain identifier, respectively. 
Incorporating domain information allows the decoder to tailor its outputs to specific domains, thereby embedding domain-specific features in the latent space during VAE training.
Since domain information is not provided to the encoder, the latent space remains general, potentially capturing shared features across domains. The procedures are:
\begin{equation}
    \hat{\mathbf{x}}_t = p_{\theta_t}(\mathbf{x}_t| \mathbf{z}_t, DomID), \qquad  \hat{\mathbf{x}}_s = p_{\theta_s}(\mathbf{x}_s| \mathbf{z}_s, DomID),\label{eq:vae-decoder}
\end{equation}
where $DomID$ is the identifier of a training domain. In our implementation, we use a one-hot encoding to convert a domain index into a binary vector. This design allows for domain-agnostic testing since we only need the encoder to generate the latent vectors.

\paragraph{\textbf{Objective function}}
The loss for training the Conditional $\beta$-VAE  includes the reconstruction loss and the Kullback-Leibler (KL) divergence. The KL divergence regularizes the learned latent posterior distributions $q_{\phi_t}(\mathbf{z}_t|\mathbf{x}_t), q_{\phi_s}(\mathbf{z}_s|\mathbf{x}_s)$, ensuring they approximate the prior distributions $p(\mathbf{z}_t), p(\mathbf{z}_s)$. The loss is written as:
\begin{align}
    L_{\text{latent}} = {}&\mathbb{E}_{\mathbf{x}}\Big [ \big ((\hat{\mathbf{x}}_t+\hat{\mathbf{x}}_s) - \mathbf{x}  \big )^2 \Big ]\\
    &{}+ \beta \Big ( \log p_{\theta_t}(\mathbf{x}_t| \mathbf{z}_t, DomID) - \text{KL} \big (q_{\phi_t}(\mathbf{z}_t|\mathbf{x}_t)||p(\mathbf{z}_t) \big ) \\
    &{}+ \log p_{\theta_s}(\mathbf{x}_s| \mathbf{z}_s, DomID) - \text{KL} \big (q_{\phi_s}(\mathbf{z}_s|\mathbf{x}_s)||p(\mathbf{z}_s) \big ) \Big ).
    \label{eq:vae-loss}
\end{align}
We introduce a weighting term, $\beta$, to regulate the capacity of the latent space in our model. Larger values of $\beta$ (e.g., $\beta > 1$) restrict the capacity of the latent vector, encouraging each dimension of the latent vector to capture distinct, conditionally independent factors from the input sequence~\cite{higgins2017beta}. $\beta$ is a hyperparameter.


\subsection{Domain Regularization}
Learning disentangled latent factors through the Conditional $\beta$-VAE may seem unintuitive for improving the generalization ability in time series forecasting. A key question arises: can we explicitly identify and use the domain-shared and domain-specific parts in the latent space to enhance generalization to unseen domains?
Doing so would not only help generalization but also improve interpretability, offering a deeper understanding of the latent vectors.
Motivated by this, we introduce \textbf{Domain Regularization}, which encourages part of the latent vectors to encode information shared across all domains, while the remaining part captures domain-specific characteristics.

Existing techniques, such as learning independent latent subspaces for domain and others~\cite{ilse2020diva}, often introduce additional parameters, increasing model complexity. Instead, we use the inherent disentangling capabilities of the $\beta$-VAE and simplify the method by partitioning the learned latent vectors into two parts: a shared part and a domain-specific part. To achieve this, we encourage the similarity of the shared parts and the dissimilarity of the domain-specific parts with a regularization loss.
We define the separation of latent vectors as follows:
\begin{align*}
 \mathbf{z}_{\text{shared}} =  \text{Concat} (\mathbf{z}_t[:index], \mathbf{z}_s[:index]), \\
  \mathbf{z}_{\text{specific}} = \text{Concat} ( \mathbf{z}_t[index:],  \mathbf{z}_s[index:]),
\end{align*}
where $index$ defines the boundary that splits the latent vector into shared and domain-specific parts. 
We introduce a hyperparameter  $0<\alpha <1$ to control this split, with the $index$ determined as $index=\alpha \times d_z$, where $d_z$ is the dimension of the latent vector. $\text{Concat}(\cdot,\cdot)$ denotes concatenation along the feature dimension.

The regularization loss term is written as follows:
\begin{equation}
\Omega(\mathbf{Z}) = \frac{1}{{N'}^2}\sum_{i_1, i_2}^{N'}|| \mathbf{z}_{\text{shared},i_1} - \mathbf{z}_{\text{shared},i_2} ||_2  -  \frac{1}{N^{\text{diff}}}\sum_{i_1, i_2; D(i_1) \neq D(i_2)}^{N'} || \mathbf{z}_{\text{specific},i_1} - \mathbf{z}_{\text{specific},i_2} ||_2, 
\label{eq:domain-reg}
\end{equation}
where $\mathbf{z}^j_{\text{shared},i_1}, \mathbf{z}^k_{\text{specific},i_1}$  denote the shared and specific latent vectors for sample $i_1$ respectively, and $D(i_1)$ indicates its domain. $N'$ denotes the total number of training samples and $N^{\text{diff}}$ denotes the number of sample pairs from different domains. 
The first term encourages the shared factors to be domain-invariant by minimizing the pairwise L2 distance between all samples. The second term enforces diversity among the domain-specific factors by maximizing differences of sample pairs from different domains.
  

\subsection{Forecasting Decoder}
After learning and disentangling the latent factors for the trend-cyclical and seasonal components, we combine them to form a unified representation for forecasting. 
We sum the two latent vectors to obtain a unified representation assuming their equal importance. Note that other combination methods could also work. 
\begin{equation}
     \mathbf{z} =  \mathbf{z}_t +  \mathbf{z}_s.
\end{equation}
Rather than relying solely on shared latent vectors for forecasting and generalization, we also incorporate domain-specific information, as they correspond to each input sample.
%
We feed the dynamically generated latent representations into a forecasting decoder, injecting common and domain-specific knowledge about temporal dependencies to improve the model's ability to generalize to unseen scenarios. We found that a linear transformation is effective in this pipeline:
\begin{equation}
  \mathbf{x}' =  \text{Concat}(\mathbf{z}, \mathbf{x})  \times  \mathbf{W}+ \mathbf{b},
  \label{eq:concat}
\end{equation}
where $\mathbf{W} \in \mathbb{R}^{(d_z + T) \times T}, \mathbf{b} \in \mathbb{R}^{T}$ are learnable parameters for the linear transformation layer. 
The enhanced input $\mathbf{x}'$, along with any possible external features $\mathbf{a}$, are fed into a forecasting decoder $\text{FcstDec}(\cdot)$, which can be any time series forecasting model.
In the probabilistic forecasting setting, we optimize the forecasting decoder by minimizing the negative log-likelihood:
\begin{equation}
    L_{\text{forecast}} = - \mathbb{E}_{\mathbf{x}} \Big [\log \big (\mathbf{y}| \text{FcstDec}(\mathbf{x}', \mathbf{a}) \big ) \Big].
    \label{eq:forecast-loss}
\end{equation}

\subsection{Training and Optimization}

We employ a two-stage training process, first training the Conditional $\beta$-VAE and then the forecasting decoder. This approach ensures the latent vectors are effectively learned in the first stage, while the second stage focuses on fully using the dynamic latent information across different domains to optimize the forecasting objective.

\paragraph{Stage 1: Pretraining the Contidional $\beta$-VAE}
We train the Contidional $\beta$-VAE components on the decomposed components $\mathbf{x}_t$ and $\mathbf{x}_s$ to learn the latent representations $\mathbf{z}_t$ and $\mathbf{z}_s$. This stage focuses on minimizing the object $L_{\text{latent}}+\Omega(\mathbf{Z})$, which combines the $\beta$-VAE loss (Eq.~\ref{eq:vae-loss}) with a domain regularization term (Eq.~\ref{eq:domain-reg}), ensuring that the latent distributions disentangle domain-shared and domain-specific factors.\footnote{We use the reparameterization trick in the training process of the VAE.}
 
\paragraph{Stage 2: Training the Forecasting Decoder and fine-tuning the Conditional $\beta$-VAE encoders}
We optimize the forecasting decoder through supervised learning on the input-output sequence pairs, minimizing the forecasting loss  $L_{\text{forecast}}$ (Eq.~\ref{eq:forecast-loss}). 
We keep the Conditional $\beta$-VAE encoders trainable, allowing them to be fine-tuned for specific tasks. 
Since domain-specific latent factors are also used for forecasting, fine-tuning the encoders helps effectively leverage domain-specific information while preventing overfitting to training domains and avoiding excessive reliance on general/shared information (e.g., in Eq.~\ref{eq:concat}).

\section{Experiments}

\subsection{Experimental Setup}
\subsubsection{Datasets} We evaluate the time series domain generalization performance using five real-world datasets from diverse fields, including web, retail, finance, and energy~\cite{deng2024domain}. The datasets we consider are: (i)~\textbf{Web-traffic}, which includes traffic data of different Wikipedia projects (e.g., en.wikipedia.org and fr.wikipedia.org);\footnote{\url{https://www.kaggle.com/competitions/web-traffic-time-series-forecasting/data}} 
(ii)~\textbf{Favorita-cat}, which contains category-level sales in the same store, treating each category as a separate domain; 
(iii)~\textbf{Favorita-store}, which comprises sales data for a single category across multiple stores;\footnote{\url{https://www.kaggle.com/competitions/favorita-grocery-sales-forecasting/data}} sale values are all log-transformed; (iv)~\textbf{Stock-volume}, which contains daily price data for different indexes tracking stock exchanges collected from Yahoo! Finance;\footnote{\url{https://www.kaggle.com/datasets/mattiuzc/stock-exchange-data/data}} and
(v)~\textbf{Power-cons}, which contains power consumption data from three different distribution networks in Tetouan City, Morocco.\footnote{\url{https://archive.ics.uci.edu/dataset/849/power+consumption+of+tetouan+city}}

A dataset summary is provided in Table~\ref{tab:real-data}, where \#Time denotes the total number of timestamps, Gran. is the Granularity, $(T, h)$ denotes the window size of the historical and predicted sequences, and $|\mathbf{a}|$ is the dimension of external features. Additional details about the datasets are in the Appendix.

\begin{table}[h]
\centering
\caption{Summary of real-world datasets.}
\label{tab:real-data}
\begin{tabular}{l l c c c c}
\toprule
\textbf{Dataset} & \textbf{Domains} &\textbf{\#Time} & \textbf{Gran.} & $(T, h)$ & |$\mathbf{a}|$\\
\midrule
{Web-traffic} & 9 projects  & 803 & day & (90, 30)  & 0     \\
{Favorita-cat} & 26 categories & 306 & day & (60, 14) & 2  \\
{Favorita-store} & 45 stores  & 306 & day & (60, 14)  & 2     \\
{Stock-volume} & 12 stocks  & 516 & day & (60, 14)  & 2     \\
Power-cons & 3 zones & 244 & hour & (120, 24) &  0\\
\bottomrule
\end{tabular}
\end{table}

\subsubsection{Evaluation metrics}
We use standard evaluation metrics to assess probabilistic forecasting performance, including both point and range accuracy~\cite{salinas2020deepar,sprangers2023parameter,deng2024domain}.
For \textit{point accuracy}, we use the normalized root mean squared error (NRMSE) and symmetric mean absolute percentage error (sMAPE)~\cite{armstrong1985crystal}.
For \textit{range accuracy}, we use the normalized quantile loss including Q(0.5) and Q(mean)~\cite{salinas2020deepar,deng2024domain}. All evaluation scores are computed and averaged on all training/test domains, with lower scores indicating better performance.

\setlength{\tabcolsep}{3.1pt} 
\begin{table*}[ht]
\small
\centering
\caption{Forecasting results of \textit{range accuracy} metrics on five real-world datasets. The method that achieves the best performance for a given base model is in \textbf{bold}.}
\label{tab:real-res}
\setlength{\tabcolsep}{2pt}
\rotatebox{0}{\begin{tabular}{l cc cc cc cc cc}
\toprule
       & \multicolumn{2}{c}{\textbf{Web-traffic}}   & \multicolumn{2}{c}{\textbf{Favorita-cat}}  & \multicolumn{2}{c}{\textbf{Favorita-store}}    & \multicolumn{2}{c}{\textbf{Stock-volume}} & \multicolumn{2}{c}{\textbf{Power-cons}} 
       \\
       \cmidrule(r){2-3}
       \cmidrule(r){4-5}
       \cmidrule(r){6-7}
       \cmidrule(r){8-9}
       \cmidrule{10-11}
        & Q(0.5)         & Q(mean)        & Q(0.5)         & Q(mean)        & Q(0.5)         & Q(mean)    & Q(0.5)         & Q(mean)     & Q(0.5)         & Q(mean)        
        \\
        \midrule
\textit{DeepAR} &  \textbf{0.155 \tiny{.030}} &  \textbf{0.122 \tiny{.018}}  & 0.122 \tiny{.040} & 0.099 \tiny{.032} & 0.021 \tiny{.001} & 0.016, \tiny{.001}  & 0.250 \tiny{.110} & 0.207 \tiny{.095} & 0.281 \tiny{.001} & 0.212 \tiny{.001}\\
+IDGM   & 0.242 \tiny{.048} & 0.191 \tiny{.034}  & 0.147 \tiny{.044} & 0.116 \tiny{.031} & 0.021 \tiny{.002} & 0.017 \tiny{.002}  & 0.271 \tiny{.084} & 0.212 \tiny{.062} & 0.297 \tiny{.021} & 0.255 \tiny{.023} \\
+Cedar  &  0.159 \tiny{.030} &  0.128 \tiny{.027}& 0.092 \tiny{.030} & \textbf{0.076 \tiny{.026}} & 0.020 \tiny{.002} & 0.016 \tiny{.001}  & 0.216 \tiny{.049} & \textbf{0.170 \tiny{.038}} & 0.281 \tiny{.001} & 0.212 \tiny{.001}\\
+\ours & 0.172 \tiny{.014} &  0.179 \tiny{.038} & \textbf{0.082} \tiny{.024} & 0.077 \tiny{.020} & \textbf{0.016 \tiny{.003}} & \textbf{0.015 \tiny{.003}} & \textbf{0.210 \tiny{.024}} & \textbf{0.170 \tiny{.029}} & \textbf{0.132 \tiny{.007}} & \textbf{0.105 \tiny{.006}} \\
\midrule
\textit{WaveNet} & 0.211 \tiny{.077}  & 0.171 \tiny{.043} & 0.121 \tiny{.031} & 0.100 \tiny{.025} & 0.021 \tiny{.001} & 0.017 \tiny{.001}  & 0.224 \tiny{.048} & 0.179 \tiny{.048} & 0.286 \tiny{.036} & 0.229 \tiny{.023} \\
+IDGM  &  0.244 \tiny{.065} &   0.190 \tiny{.048}  &  0.151 \tiny{.043} & 0.124 \tiny{.036} & 0.041 \tiny{.019} & 0.034 \tiny{.017}   & 0.276 \tiny{.108} & 0.224 \tiny{.098} & 0.257 \tiny{.028} & 0.224 \tiny{.026} \\
+Cedar  &  0.175 \tiny{.039} & 0.140 \tiny{.027}  & 0.088 \tiny{.027} & \textbf{0.073 \tiny{.023}} & 0.017 \tiny{.001} & \textbf{0.013 \tiny{.001}}   & 0.232 \tiny{.055} & 0.183 \tiny{.046} & 0.255 \tiny{.031} & 0.203 \tiny{.034}\\
+\ours  & \textbf{0.137 \tiny{.024}} & \textbf{0.120 \tiny{.013}}  &  \textbf{0.082 \tiny{.021}} & 0.077 \tiny{.016} & \textbf{0.015 \tiny{.001}} & 0.014 \tiny{.004} & \textbf{0.217 \tiny{.037}} & \textbf{0.173 \tiny{.035}} & \textbf{0.161 \tiny{.062}} & \textbf{0.133 \tiny{.055}} \\
\midrule
\textit{DLinear} & \textbf{0.152 \tiny{.003}} & \textbf{0.119 \tiny{.008}} & \textbf{0.088 \tiny{.021}} & \textbf{0.075 \tiny{.016}} & \textbf{0.017 \tiny{.002}} & 0.014 \tiny{.002} & \textbf{0.209 \tiny{.040}} &  0.206 \tiny{.074} & 0.224 \tiny{.017} & 0.189 \tiny{.018}   \\
+IDGM   &  0.161 \tiny{.009} & 0.140 \tiny{.019} &  0.091 \tiny{.025} & 0.079 \tiny{.018} & 0.018 \tiny{.001} & 0.014 \tiny{.001}  & 0.212 \tiny{.042} &  0.218 \tiny{.053} & 0.225 \tiny{.024} & 0.181 \tiny{.021} \\
+\ours  & 0.172 \tiny{.032} & 0.145 \tiny{.040} & \textbf{0.088 \tiny{.021} }& \textbf{0.075 \tiny{.017}} & \textbf{0.017 \tiny{.001} }& \textbf{0.013 \tiny{.001}}  &  0.215 \tiny{.044} & \textbf{0.179 \tiny{.046}} & \textbf{0.212 \tiny{.016}} & \textbf{0.172 \tiny{.010}} \\
\bottomrule
\end{tabular}}\label{tab:res-range}
\end{table*}

\subsubsection{Baselines}
We compare our proposed method \ours~ with state-of-the-art domain generalization methods easily applied in time series forecasting: a gradient matching approach \textbf{IDGM}~\cite{shi2021gradient} and the cross-domain regularization method with difficulty awareness, \textbf{Cedar}~\cite{deng2024domain}.\footnote{We exclude weaker domain generalization methods such as DANN~\cite{ganin2016domain} and wERM due to their inferior generalization performance in time series forecasting~\cite{deng2024domain}.} 
We also compare our method against the recent LLM-based cross-domain learning model, \textbf{UniTime}~\cite{liu2024unitime}.

For the forecasting decoder, we use the RNN-based \textbf{DeepAR}~\cite{salinas2020deepar}, CNN-based \textbf{WaveNet}~\cite{oord2016wavenet} following a recent study~\cite{deng2024domain}. We also include an MLP-based \textbf{DLinear}~\cite{zeng2023transformers} and a lightweight, effective LLM-based model, \textbf{GPT4TS}~\cite{zhou2023one}. Since \textbf{GPT4TS} and \textbf{UniTime} are inherently for point forecasting, we use the mean squared loss (MSE) as the forecasting loss and only report their \textit{point accuracy} results. 
We do not apply the Cedar baseline to DLinear because DLinear does not generate intermediate hidden features that can be used for feature alignment.
All implementation details are provided in the Appendix. 

\subsection{Main Results}
\setlength{\tabcolsep}{3.1pt} 

\begin{table*}[t]
\small
\centering
\caption{Forecasting results of \textit{point accuracy} metrics on five real-world datasets. The method that achieves the best performance for a given base model is in \textbf{bold}.}
\label{tab:real-res-point}
\rotatebox{0}{\begin{tabular}{l cc cc cc cc cc}
\toprule
       & \multicolumn{2}{c}{\textbf{Web-traffic}}  & \multicolumn{2}{c}{\textbf{Favorita-cat}}  & \multicolumn{2}{c}{\textbf{Favorita-store}}     & \multicolumn{2}{c}{\textbf{Stock-volume}} & \multicolumn{2}{c}{\textbf{Power-cons}} 
       \\
       \cmidrule(r){2-3}
       \cmidrule(r){4-5}
       \cmidrule(r){6-7}
       \cmidrule(r){8-9}
       \cmidrule(r){10-11}
          & NRMSE          & sMAPE          & NRMSE          & sMAPE          & NRMSE          & sMAPE          & NRMSE           & sMAPE     & NRMSE           & sMAPE          \\
          \midrule
UniTime  & 0.225 \tiny{.039} & 0.145 \tiny{.020}  &  0.139 \tiny{.047} & 0.197 \tiny{.067} & 0.036 \tiny{.003} & 0.029 \tiny{.003} & 0.524 \tiny{.093} & 0.213 \tiny{.024} & 0.345 \tiny{.014} & 0.293 \tiny{.015}\\
        \midrule
\textit{DeepAR}  & \textbf{0.224 \tiny{.055}}  & \textbf{0.157 \tiny{.015}}   & 0.175 \tiny{.066}  & 0.218 \tiny{.064} & 0.027 \tiny{.002}  & 0.021 \tiny{.001}   & 0.534 \tiny{.193}  & 0.639 \tiny{.279} &0.340 \tiny{.002} & 0.296 \tiny{.001} \\
+IDGM    & 0.312 \tiny{.076}  &  0.264 \tiny{.072} & 0.208 \tiny{.065}  & 0.241 \tiny{.053} & 0.027 \tiny{.002}  & 0.021 \tiny{.002}   & 0.547 \tiny{.139}  & 0.703 \tiny{.293} & 0.370 \tiny{.033} & 0.312 \tiny{.025} \\
+Cedar   & 0.226 \tiny{.057}  &  0.177 \tiny{.070}  & 0.136 \tiny{.050}  & 0.193 \tiny{.070} & 0.026 \tiny{.002}  & 0.020 \tiny{.002}   & \textbf{0.482 \tiny{.121}}  & 0.295 \tiny{.128} & 0.339 \tiny{.002} & 0.296 \tiny{.001} \\
+\ours  & 0.243 \tiny{.017} & 0.172 \tiny{.026}  & \textbf{0.122 \tiny{.032}} & \textbf{0.188 \tiny{.069}} & \textbf{0.022 \tiny{.003}} & \textbf{0.016 \tiny{.003}} & 0.484 \tiny{.096} & \textbf{0.234 \tiny{.031}} &  \textbf{0.165 \tiny{.007}} & \textbf{0.146 \tiny{.006}} \\

\midrule
\textit{WaveNet} & 0.287 \tiny{.114}  & 0.222 \tiny{.095}   & 0.172 \tiny{.053}  & 0.218 \tiny{.057} & 0.027 \tiny{.002}   & 0.021 \tiny{.001}  & 0.482 \tiny{.097}  & 0.541 \tiny{.211} & 0.352 \tiny{.038} & 0.296 \tiny{.034} \\
+IDGM     & 0.333 \tiny{.100}  &  0.231 \tiny{.045}  & 0.211 \tiny{.073}  & 0.244 \tiny{.056} & 0.033 \tiny{.005}   & 0.025 \tiny{.004}  & 0.555 \tiny{.180}  & 0.548 \tiny{.205} & 0.333 \tiny{.041} & 0.263 \tiny{.022}\\
+Cedar   & 0.237 \tiny{.064} & 0.191 \tiny{.057}   & 0.130 \tiny{.045}  & 0.193 \tiny{.068} & 0.022 \tiny{.002}   & 0.016 \tiny{.001}   & \textbf{0.481 \tiny{.110}} & 0.583 \tiny{.286} & 0.315 \tiny{.030} & 0.266 \tiny{.032}\\
+\ours  & \textbf{0.209 \tiny{.048}} & \textbf{0.140} \tiny{.033}  & \textbf{0.124 \tiny{.030}} & \textbf{0.191 \tiny{.068}} & \textbf{0.021 \tiny{.002}} & \textbf{0.015 \tiny{.001}}& 0.492 \tiny{.110} & \textbf{0.243 \tiny{.036}} & \textbf{0.204 \tiny{.074}} & \textbf{0.177 \tiny{.060}} \\

\midrule
\textit{DLinear} & \textbf{0.233 \tiny{.018}} & 0.170 \tiny{.022} & \textbf{0.127 \tiny{.030}} & 0.195 \tiny{.066} & 0.023 \tiny{.002} & \textbf{0.017 \tiny{.002}}  & \textbf{0.475 \tiny{.107}} & 0.369 \tiny{.198} & 0.304 \tiny{.029} & 0.235 \tiny{.018}\\
+IDGM & 0.221 \tiny{.015} &  \textbf{0.169 \tiny{.026}} & 0.130 \tiny{.035} & \textbf{0.194 \tiny{.067}} & 0.023 \tiny{.001} & 0.018 \tiny{.001}  & 0.481 \tiny{.114} & \textbf{0.341 \tiny{.125}} & 0.346 \tiny{.076} & 0.233 \tiny{.031} \\
+\ours & 0.238 \tiny{.043} & 0.181 \tiny{.048} & \textbf{0.127 \tiny{.030}} & \textbf{0.194 \tiny{.066}} & \textbf{0.022 \tiny{.001}} & \textbf{0.017 \tiny{.001}} & 0.478 \tiny{.112} & 0.344 \tiny{.140} & \textbf{0.267 \tiny{.021}} & \textbf{0.214 \tiny{.014}} \\

\midrule

\textit{GPT4TS} &  0.380 \tiny{.200} & 0.178 \tiny{.046}  & 0.130 \tiny{.038} & 0.190 \tiny{.068} & \textbf{0.025 \tiny{.003}} & \textbf{0.018 \tiny{.002}} & 0.510 \tiny{.126} & 0.203 \tiny{.027} & 0.342 \tiny{.016} & 0.280 \tiny{.014}  \\
+IDGM &  0.405 \tiny{.202} & 0.195 \tiny{.058} & \textbf{0.129 \tiny{.039}} &  \textbf{0.189 \tiny{.066}} & \textbf{0.025 \tiny{.004}} & \textbf{0.018 \tiny{.003}}  & 0.528 \tiny{.159} & 0.220 \tiny{.032} & 0.316 \tiny{.016}  & 0.258 \tiny{.021} \\
+Cedar & 0.354 \tiny{.183} & 0.185 \tiny{.057} & 0.142 \tiny{.055} & 0.194 \tiny{.066} & \textbf{0.025 \tiny{.003}} & \textbf{0.018 \tiny{.002}}  & 0.510 \tiny{.126} & 0.203 \tiny{.027} & 0.336 \tiny{.011} & 0.282 \tiny{.014}\\
+\ours & \textbf{0.237 \tiny{.075}} & \textbf{0.154 \tiny{.036}}  & \textbf{0.129 \tiny{.077}} & 0.195 \tiny{.070} & \textbf{0.025 \tiny{.003}} & 0.019 \tiny{.002} & \textbf{0.476 \tiny{.104}} & \textbf{0.193 \tiny{.024}} & \textbf{0.255 \tiny{.030}} & \textbf{0.202 \tiny{.027}} \\

\bottomrule
\end{tabular}}\label{tab:res-point}
\end{table*} 
Tables~\ref{tab:res-range} and~\ref{tab:res-point} present range and point accuracy results, respectively. We apply our method \ours{} to DeepAR, WaveNet, DLinear, and GPT4TS, comparing it with other generalization techniques. 

The \textit{range accuracy} results presented in Table~\ref{tab:res-range} demonstrate that \ours{} achieves the overall best performance. In many cases, \ours{} attains higher Q(0.5) values more easily than Q(mean) compared with baselines (e.g., on the Favorita-cat and Favorita-store datasets). Here, Q(0.5) refers to the normalized quantile loss at quantile 0.5, while Q(mean) represents the average normalized quantile loss across a range of quantiles from 0.1 to 0.9. Q(0.5) represents the median of the target distribution. A lower Q(0.5) suggests that a model effectively predicts the central tendency of the data. This phenomenon suggests our model is proficient at estimating the trends of future movements in new domain data. 
We observe that on the Web-traffic datasets, base models like DeepAR and DLinear achieve the best performance. However, \ours{} applied to WaveNet achieves the lowest Q(0.5). This suggests that while base models can limit the effectiveness of generalization enhancements, our approach can largely improve generalization when combined with a proper base model. 
We found that DLinear is effective sometimes even though it is quite simple, with only linear layers. We believe the limited effectiveness of \ours{} on DLinear stems from its linear focus on directly modeling trend and seasonal components, where incorporating additional knowledge may not provide significant benefits. Moreover, our qualitative analysis indicates that the generated sequences of DLinear fail to capture the true trend and instead exhibit a lot of noise (see Sec.~\ref{sec:showcases}).

Table~\ref{tab:res-point} presents the \textit{point accuracy} results in NRMSE and sMAPE, where \ours{} outperforms its counterparts in most cases.  An interesting finding is that generalization techniques appear less effective on GPT4TS, when GPT4TS already achieves strong performance, e.g., in the Favorita-cat and Favorita-store datasets. This may be due to the limited impact of generalization add-ons on pretrained LLMs, as these modifications typically adjust or affect only a small number of parameters within the overall model. The performance differences across models are relatively small on the Favorita-cat and Favorita-store datasets. This could be due to the log transformation~\cite{sprangers2023parameter,deng2024domain} stabilizing the data.

For the cross-domain learning model UniTime~\cite{liu2024unitime}, it performs well on the Web-traffic dataset but may not be a strong baseline for other datasets. This is because UniTime focuses on adapting across broader application domains, such as forecasting hourly traffic and daily exchange rates, using sentences about domain knowledge to guide the training of a pretrained LLM. In contrast, our work addresses more fine grained domains (e.g., different websites), where similar instructional sentences may not provide benefits. The key challenge instead lies in effectively capturing the common knowledge across training domains.

\subsection{Ablation Studies} 
To verify the effectiveness of our proposed model, we conduct ablation studies by altering the main components in our design individually. We consider the following variants: (1) \ours(e2e) modifies the two-stage training into an end-to-end training procedure. (2) \ours(w/o Reg) removes the domain regularization loss. (3) \ours(w/o deC) removes the time series decomposition and uses a single Conditional $\beta$-VAE to model the unified latent representation $\mathbf{z}$. (4) \ours(w/o Spe) feeds only the domain shared latent vectors $\mathbf{z}_{\text{shared}}$ to the forecasting decoder. (5) \ours(w/o Cond) removes \textit{DomID} from the input of the conditional decoder (Eq.~\ref{eq:vae-decoder}).

We follow the same settings as in our main experiments and report the results on four datasets using two base models in Table~\ref{tab:ablation}. 
We observe that the full model achieves the best performance on Stock-volume.
\ours(w/o Spe) is a key variant that uses only shared latent factors for forecasting, achieving the best performance on the Web-traffic dataset, sometimes outperforming the best baseline. 
This approach seems more intuitive, yet it might not generalize well in scenarios where domain shifts are insignificant. In that case, estimating domain-specific nuances is important to improve performance within specific domains.
We leave it to future work to explore strategies that dynamically maximize the benefits of shared latent factors while retaining key domain-specific insights on diverse datasets.

For other variants, end-to-end training of the entire framework, i.e., \ours(e2e), does not yield better results, highlighting the importance of first learning the latent factors effectively before training the forecasting model.
We observe that simpler variants can achieve outstanding performance.
For instance, on the Web-traffic and Stock-volume datasets, \ours{} (w/o deC) effectively captures latent factors without time series decomposition when using the DeepAR model. This may be because the datasets have relatively simple patterns that are easy to capture. Time series visualizations of the datasets are shown in later sections. Meanwhile, this variant reduces the parameters of the full framework (i.e., requiring only one conditional $\beta$-VAE), helping to prevent overfitting. 

\setlength{\tabcolsep}{3.1pt} 

\begin{table}[!h]
\small
\centering
\caption{Forecasting results in Q(0.5) and sMAPE with variations of \ours{} based on DeepAR and DLinear. The best performances are highlighted in bold.}
\label{tab:real-res}
\rotatebox{0}{\begin{tabular}{c l cc cc}
\toprule
    &  \multirow{2}[1]{*}{Model}   & \multicolumn{2}{c}{\textbf{Web-traffic}}  & \multicolumn{2}{c}{\textbf{Stock-volume}}   
    \\
    \cmidrule(r){3-4}
    \cmidrule{5-6}
     &   & Q(0.5)   & sMAPE      & Q(0.5)  &  sMAPE      \\
        \midrule
\multirow{4}[1]{*}{\rotatebox{90}{\textit{DeepAR}}}  & +\ours  & 0.1717 \tiny{.0142} & 0.1724 \tiny{.0255}  & \textbf{0.2100 \tiny{.0236}} & \textbf{0.2335 \tiny{.0311}}\\
& +\ours(e2e) & 0.1713 \tiny{.0144} & 0.1722 \tiny{.0259} &  0.2162 \tiny{.0258} &  0.2375 \tiny{.0330}  \\
& +\ours(w/o Reg) & 0.1718 \tiny{.0141} & 0.1725 \tiny{.0256} &  0.2167 \tiny{.0265} & 0.2426 \tiny{.0311}\\
& +\ours(w/o deC) & 0.1497 \tiny{.0291} & 0.1577 \tiny{.0418} & 0.2113 \tiny{.0249} & 0.2366 \tiny{.0316}  \\
& +\ours(w/o Spe)& \textbf{0.1494 \tiny{.0254}} & \textbf{0.1572 \tiny{.0405}} &0.2135 \tiny{.0260} & 0.2367 \tiny{.0304}\\ 
   & +\ours(w/o Cond) & 0.1635 \tiny{.0172} & 0.1668 \tiny{.0301} & 0.2130 \tiny{.0317} & 0.2445 \tiny{.0289}\\
\midrule
\multirow{4}[1]{*}{\rotatebox{90}{\textit{DLinear}}}  & +\ours & 0.1724 \tiny{.0317} & 0.1811 \tiny{.0479}  & \textbf{0.2145 \tiny{.0441}}  & \textbf{0.3441 \tiny{.1400}} \\
  & +\ours(e2e) & 0.1756 \tiny{.0277} & 0.1821 \tiny{.0432} & 0.2213 \tiny{.0461} & 0.3828 \tiny{.1607}  \\
  & +\ours(w/o Reg) & 0.1789 \tiny{.0203} & 0.1874 \tiny{.0382} & 0.2170 \tiny{.0442} & 0.3426 \tiny{.1333}\\
  & +\ours(w/o deC) & 0.1747 \tiny{.0178} & 0.1790 \tiny{.0291} & 0.2168 \tiny{.0451} &  0.3873 \tiny{.1336} \\
& +\ours(w/o Spe) &  \textbf{0.1669 \tiny{.0184}} & \textbf{0.1717 \tiny{.0283}} & 0.2156 \tiny{.0456} & 0.3610 \tiny{.1560} \\
   & +\ours(w/o Cond) &  0.1773 \tiny{.0171} & 0.1819 \tiny{.0308} & 0.2203 \tiny{.0486} & 0.3843 \tiny{.1800} \\


\bottomrule
\end{tabular}}\label{tab:ablation}
\end{table}

\subsection{Model Analysis}
We conduct several analyses to assess various aspects of our approach and present the main findings in this section. Additional analyses and supporting evidence are provided in the Appendix.

  \begin{figure*}[ht!]
  \centering 
  \begin{subfigure}{.195\textwidth}
    \centering
\includegraphics[width=1.0\linewidth]{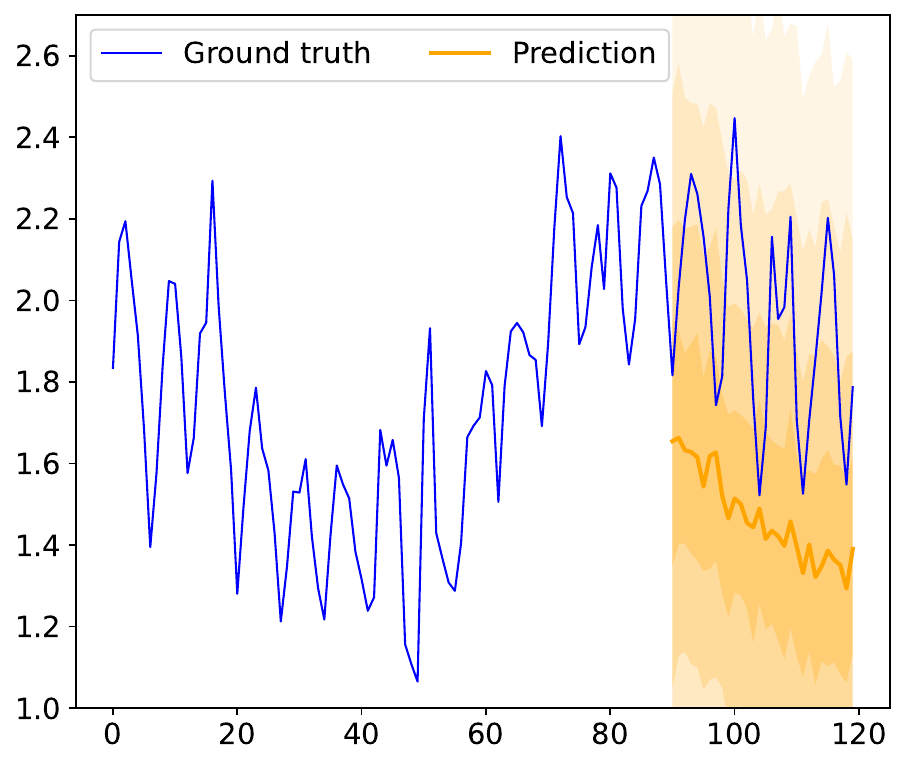}\caption{WaveNet}
  \end{subfigure}
    \begin{subfigure}{.195\textwidth}
    \centering
\includegraphics[width=1\linewidth]{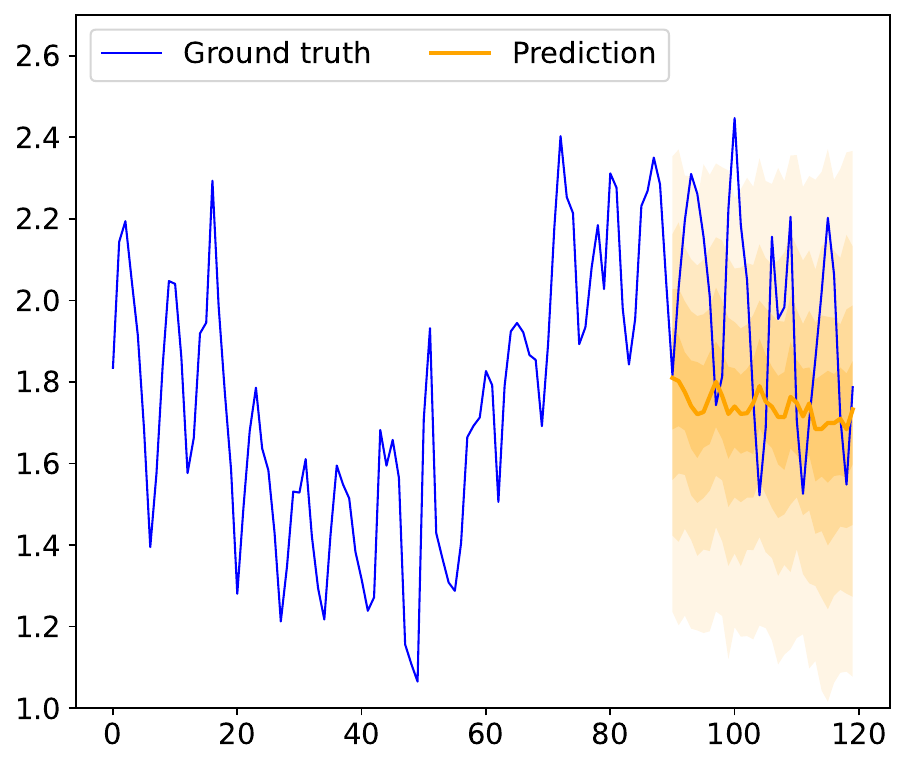}\caption{WaveNet+IDGM}
  \end{subfigure} 
    \begin{subfigure}{.195\textwidth}
    \centering
    \includegraphics[width=1.0\linewidth]{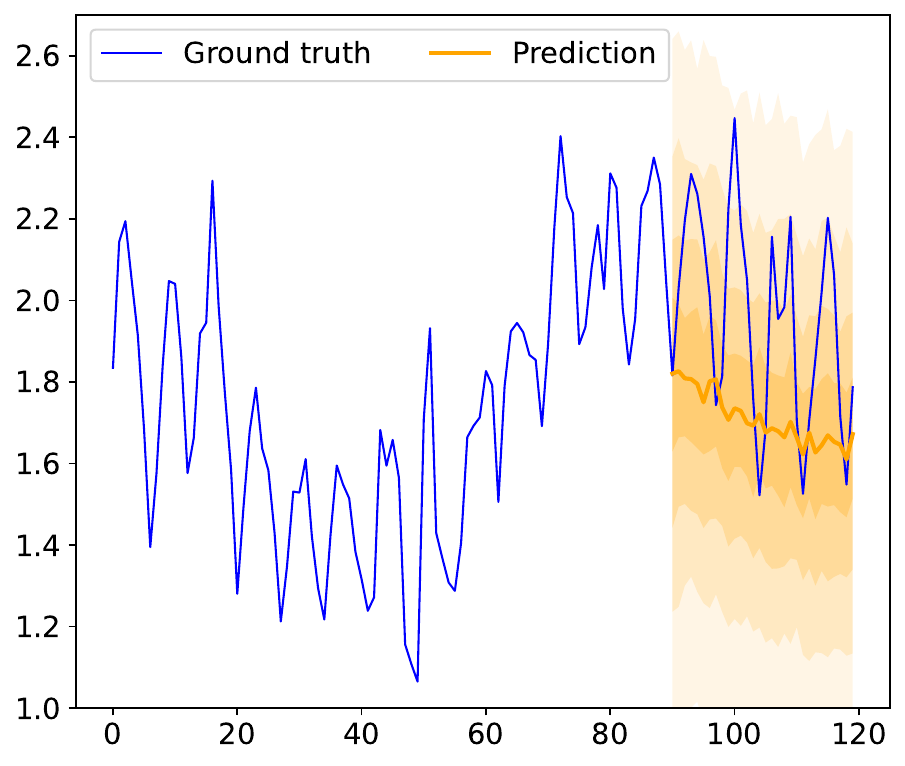}\caption{WaveNet+Cedar}
  \end{subfigure} 
  \begin{subfigure}{.195\textwidth}
    \centering
\includegraphics[width=1.0\linewidth]{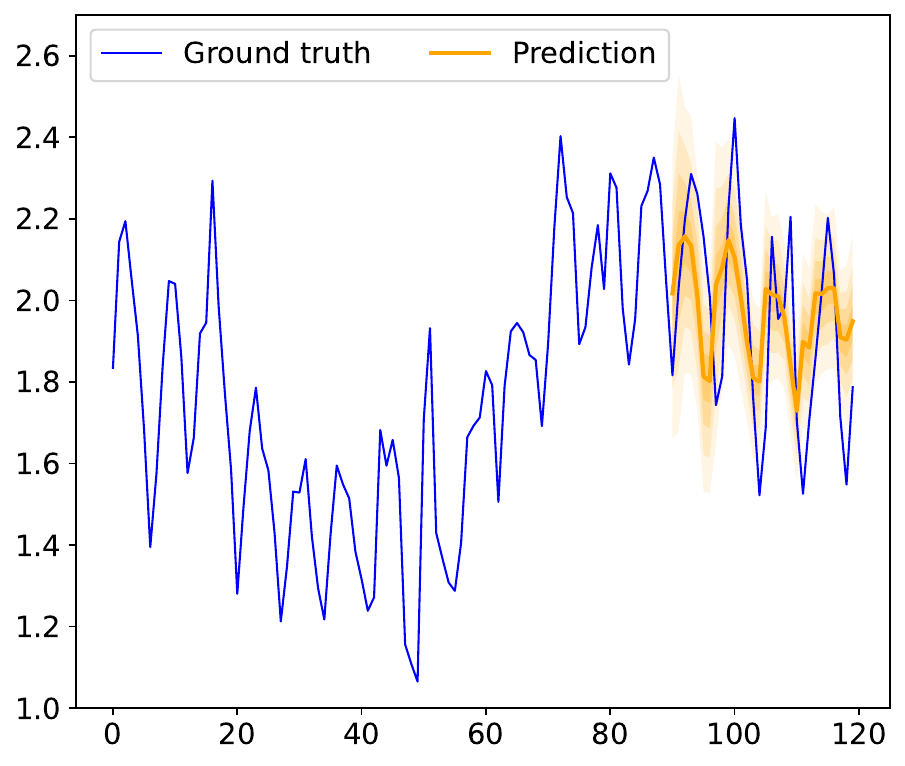}\caption{WaveNet+\ours}
  \end{subfigure} 
    \begin{subfigure}{.195\textwidth}
    \centering
\includegraphics[width=1.0\linewidth]{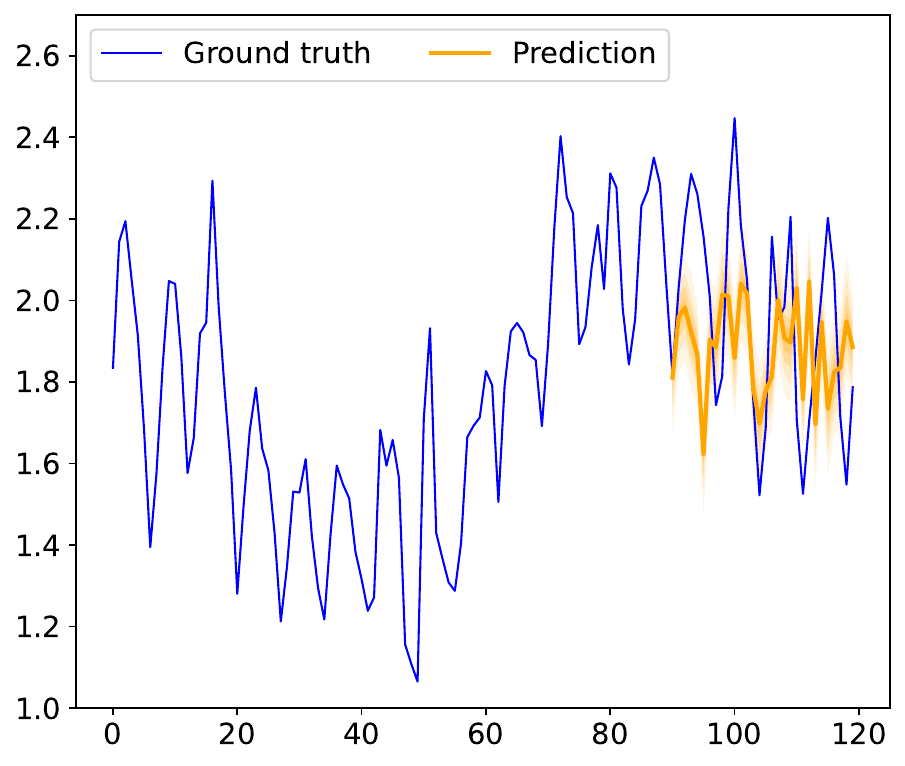}\caption{DLinear}
  \end{subfigure} 
 \caption{Forecasting results for a test domain sample from the Web-traffic dataset, using DLinear and WaveNet with various generalization methods.} 
  \label{fig:showcases}
 \end{figure*}
 
\subsubsection{Latent space analysis} 
\begin{figure}[t]
  \centering 
    \begin{subfigure}{.4\textwidth}
    \centering
    \includegraphics[width=1.0\linewidth]{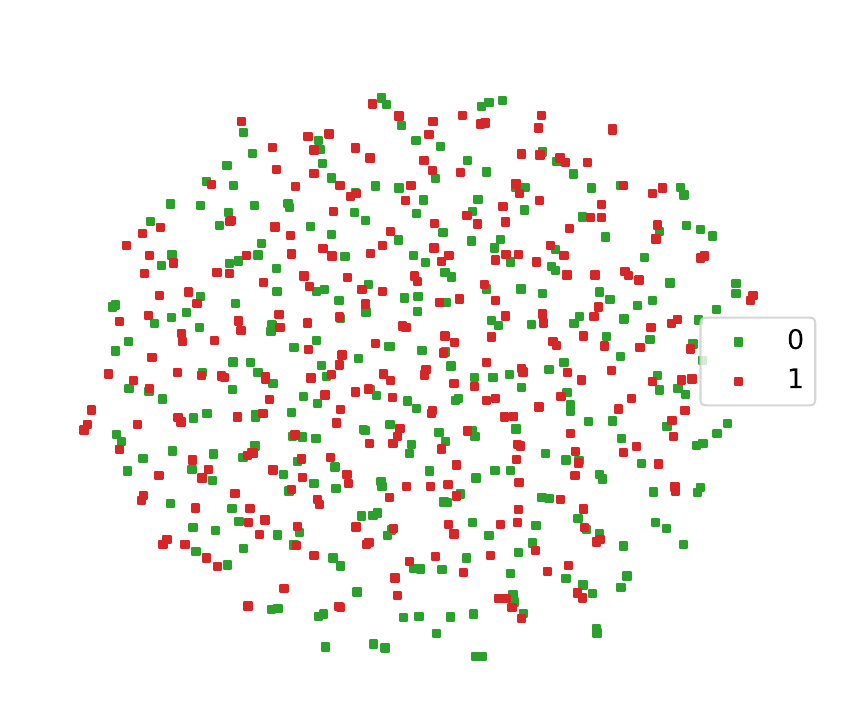}
  \end{subfigure} 
     \hspace{10pt}
  \begin{subfigure}{.36\textwidth}
    \centering
\includegraphics[width=1.0\linewidth]{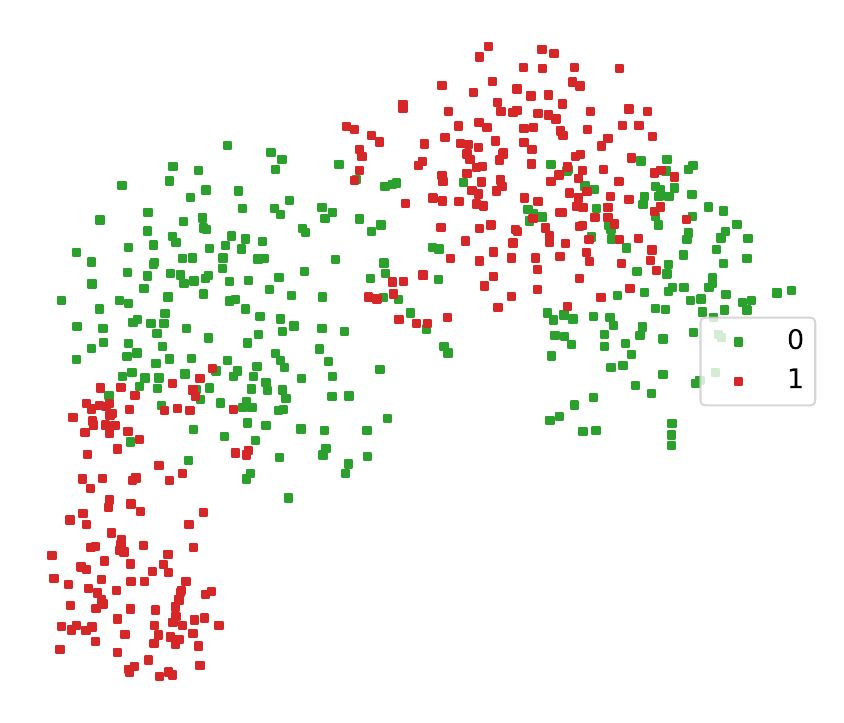}
  \end{subfigure} 

  \begin{subfigure}{.4\textwidth}
    \centering
    \includegraphics[width=1.0\linewidth]{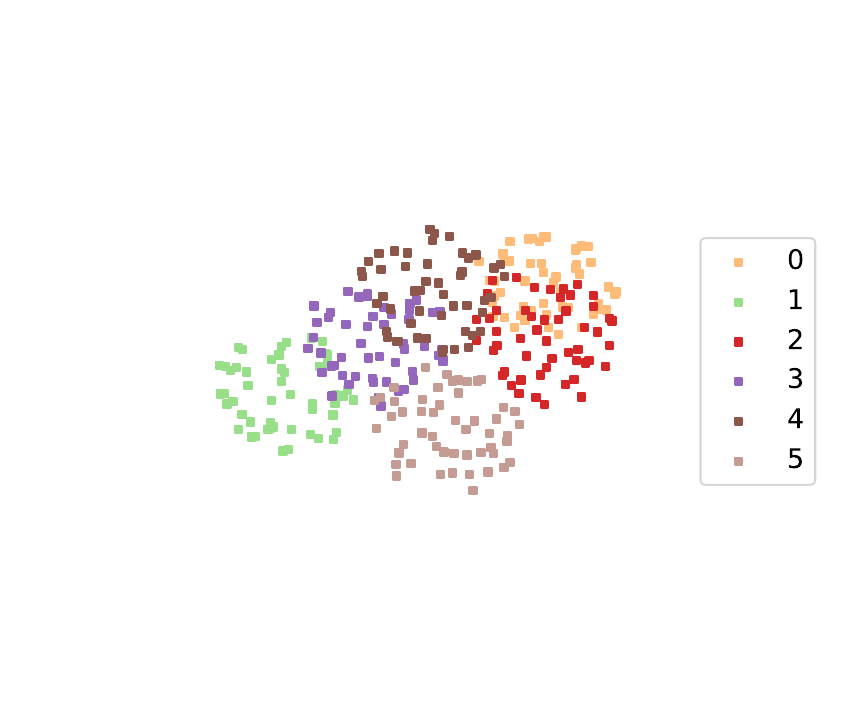}\caption{Domain-shared component}
  \end{subfigure} 
   \hspace{10pt}
  \begin{subfigure}{.36\textwidth}
    \centering
\includegraphics[width=1.0\linewidth]{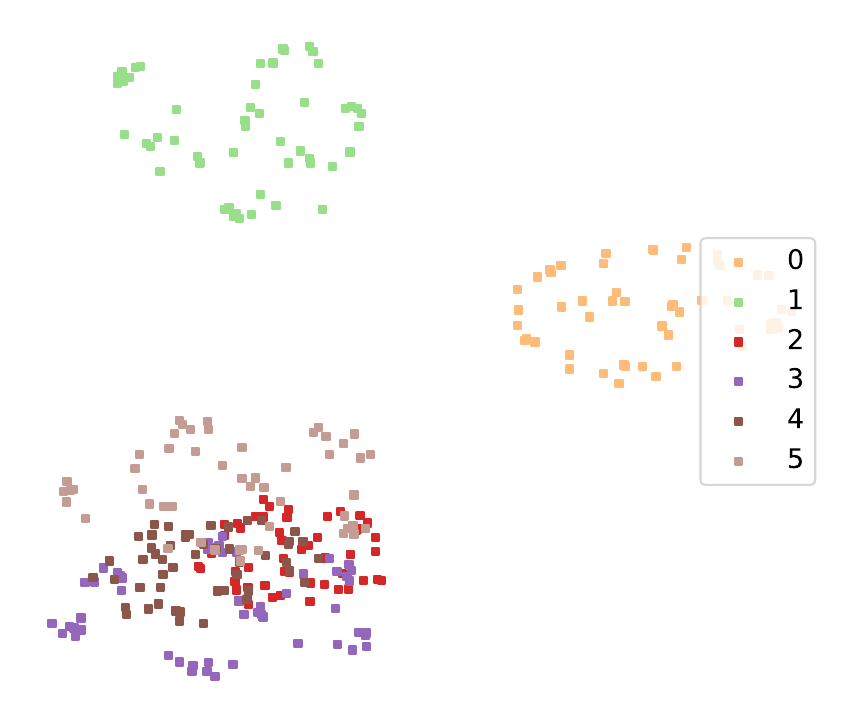}\caption{Domain-specific component}
  \end{subfigure}  
 \caption{Latent space t-SNE visualization of the domain-shared and domain-specific components on the test domains of the Web-traffic (top) and Favorita-cat (bottom) datasets.} 
  \label{fig:latent-analysis}
 \end{figure}

We analyze the learned latent factors to assess whether the domain-shared and domain-specific components exhibit distinct distributions. In Figure~\ref{fig:latent-analysis}, we visualize the latent factors using t-SNE~\cite{van2008visualizing} for both domain-shared and domain-specific parts on test domains from different datasets. 
We use combined embeddings from our model applied with DeepAR for the Favorita-cat and WaveNet for the Web-traffic dataset given their superior performance. The results show a clear overlap in the distributions of the domain-shared components across different domains, while the domain-specific parts are more distinct and separable. 
This demonstrates that our approach effectively captures shared patterns while also identifying domain-specific patterns in unseen domains.

\subsubsection{Forecast horizon sensitivity} We evaluate the performance of our approach in long-term forecasting by varying the forecast window size from 30 to 60 on the Web-traffic dataset. Figure~\ref{fig:longterm-analysis} shows the Q(0.5) results for different models. We observe that DeepAR, combined with \ours{}, maintains consistently strong performance. It highlights the strength of our model in capturing information that enhances generalization even for longer forecast horizons.
DLinear is a strong base model but performs worse than our models in long-term forecasting.

\subsubsection{Computational analysis} 
\begin{figure}[h]
  \centering 
      \begin{subfigure}{.4\textwidth}
    \centering
\includegraphics[width=1.0\linewidth]{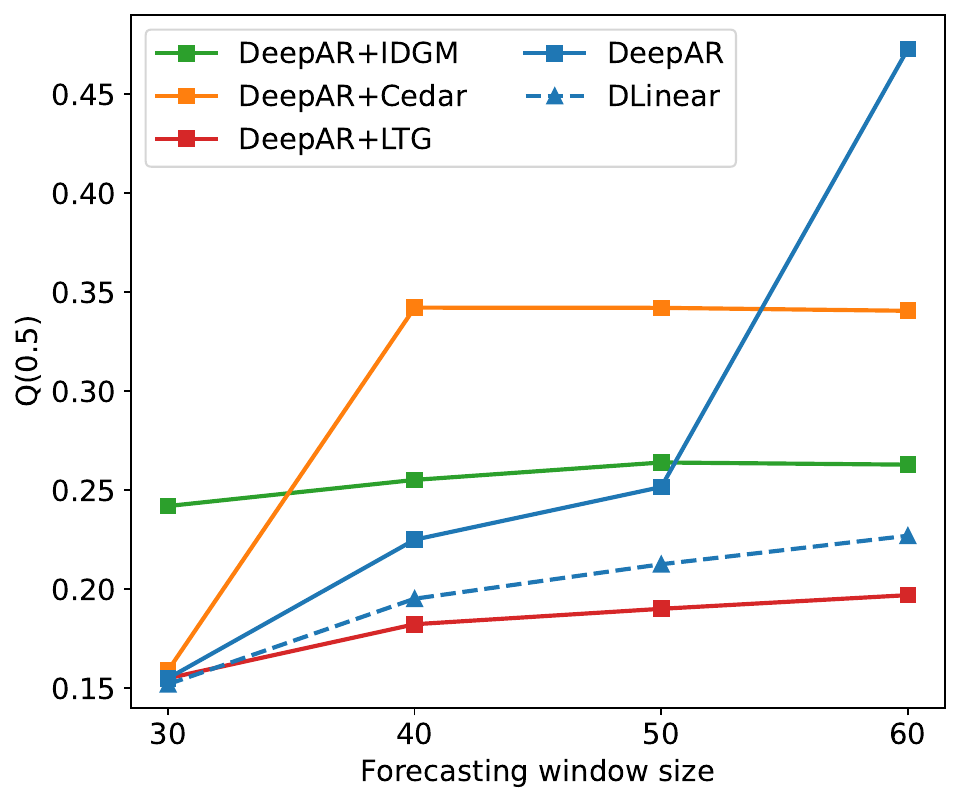}\caption{Forecasting window analysis}\label{fig:longterm-analysis}
  \end{subfigure} 
   \hspace{10pt}
  \begin{subfigure}{.4\textwidth}
    \centering
\includegraphics[width=1.0\linewidth]{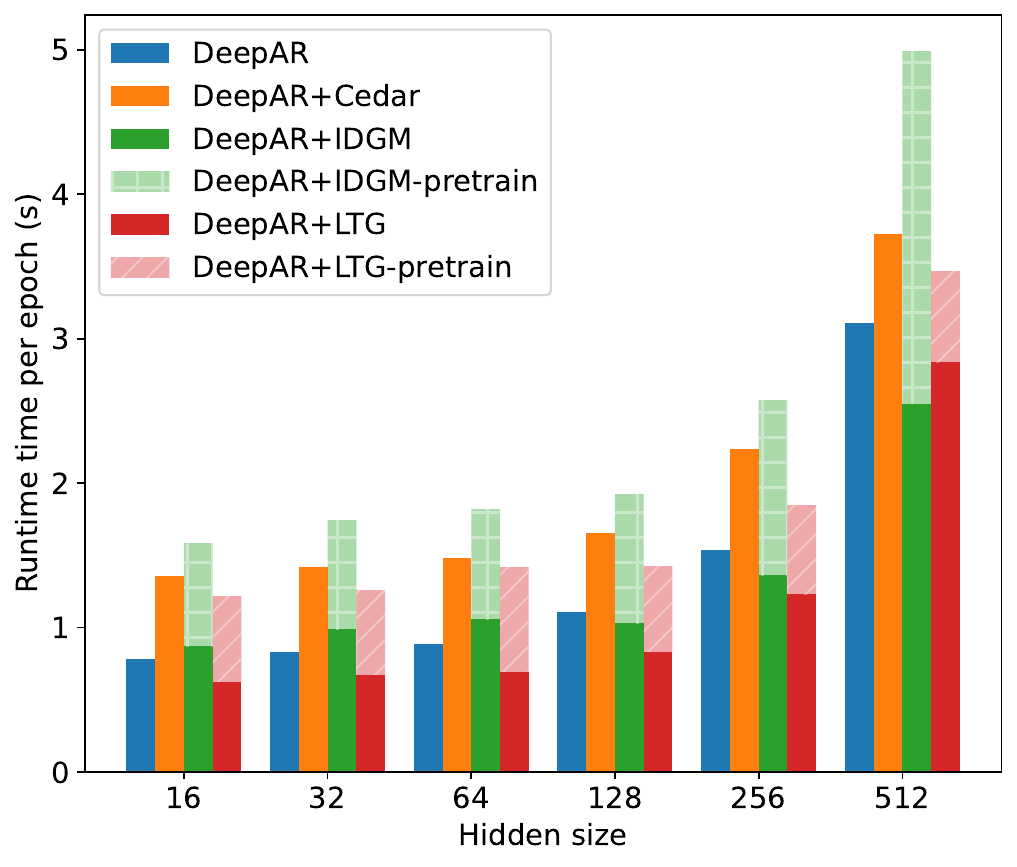}\caption{Runtime analysis}\label{fig:time-analysis}
  \end{subfigure}  
 \caption{Forecasting window sensitivity and runtime analysis on the Web-traffic dataset using DeepAR as the base model.} 
 \end{figure}

We evaluate the computational efficiency of \ours{} on the Web-traffic dataset.
We measure the average runtime per epoch for different generalization methods based on DeepAR. We also consider variations in model size (i.e., hidden size), as larger models are often preferred for handling larger and more complex datasets. We report the runtime for both training and pretraining in Figure~\ref{fig:time-analysis}. 
%
We observe that \ours{} consistently requires less total training time than Cedar, and in some cases, its training time is even lower than that of the base model.
Meanwhile, compared to the gradient matching baseline IDGM -- which also includes a pretraining process -- our method requires significantly less time for pretraining. Additionally, as the hidden size increases, the pretraining time for \ours{} scales almost linearly, while it increases more rapidly for IDGM.  
It is worth noting that in experiments if the pretraining stage of \ours{} takes more epochs, it typically requires fewer epochs during the training phase.
This indicates that it adds a very limited total computational load to the base model.

\subsubsection{Showcases}\label{sec:showcases}
Figure~\ref{fig:showcases} presents the quantile forecasting results of different generalization techniques based on WaveNet on the Web-traffic dataset. It is clear that \ours{} provides more accurate estimates of both trend and seasonality compared to others on unseen datasets. We also include the forecasting results of DLinear, given its strong quantitative performance on this dataset. While DLinear effectively captures the trend, its predictions exhibit more noise, with greater fluctuations that resemble random guessing.

 \section{Conclusion}
We present a novel framework for domain generalization in time series forecasting by mining latent factors for temporal dependencies within and across domains, which serve as crucial knowledge for forecasting in unseen domains. We introduce a conditional $\beta$-VAE to capture disentangled latent factors incorporating domain information. A decomposition module is employed to break down time series into two components, enhancing the modeling of latent factors. We further introduce domain regularization to distinguish between domain-shared and domain-specific parts of the latent space, promoting disentanglement and providing insights for downstream tasks. 
We conduct extensive experiments to evaluate the effectiveness of our approach.

A potential limitation of this method is that the latent vectors, implying temporal dependencies, are used only as additional input knowledge. 
Adding prior knowledge at the input may not always improve generalization and can be constrained by the base model. Future work will explore methods to use latent factors to regulate the training of time series forecasting models.

\section*{Acknowledgements}
This research was (partially) funded by 
Ahold Delhaize, 
the Hybrid Intelligence Center, a 10-year program funded by the Dutch Ministry of Education, Culture and Science through the Netherlands
Organisation for Scientific Research, https://hybrid-intelligence-centre.nl, 
project LESSEN with project number NWA.1389.20.183 of the research program NWA ORC 2020/21, which is (partly) financed by the Dutch Research Council (NWO),
project ROBUST with project number KICH3.LTP.20.006, which is (partly) financed by the Dutch Research Council (NWO) and the Dutch Ministry of Economic Affairs and Climate Policy (EZK) under the program LTP KIC 2020-2023,
and
the FINDHR (Fairness and Intersectional Non-Discrimination in Human Recommendation) project that received funding from the European Union's Horizon Europe research and innovation program under grant agreement No 101070212.

All content represents the opinion of the authors, which is not necessarily shared or endorsed by their respective employers and/or sponsors.

\bibliographystyle{abbrvnat}  
\bibliography{references}

\appendix

\section{Ethical Considerations} 

The research topic of this paper, domain generalization in time series forecasting, can impact ML applications in several ways. It enhances the domain generalization capability of time series forecasting models by mining latent factors, leading to more accurate predictions in new situations. This improvement positively influences decision-making and fairness across various industries, including web, retail, finance, and energy. By tackling the domain shift challenges, this research also fosters the development of more robust and adaptable AI systems. However, it is crucial to address underlying ethical concerns, including fairness and bias, to ensure equitable distribution of these advancements for the benefit of society as a whole.

\section{Datasets}
We use five real-world datasets for evaluation. Three datasets were introduced in prior work~\cite{deng2024domain}, in the areas of retail and finance.
We use the provided data and code to load and process the data.
We also introduce two new datasets in the web and energy domains:
\begin{itemize}[leftmargin=*]
    \item \textbf{Web-traffic}. This dataset contains daily traffic of different Wiki\-pedia projects (e.g., en.wikipedia.org and fr.wikipedia.org).\footnotemark{} We use data from 2015-07-01 to 2016-04-15 for training, 2016-04-15 to 2016-07-01 for validation, and the remainder until 2017-09-10 for testing. Missing values are filled with 0, and a simple preprocessing step scales the data by dividing all values by 1e7.
    \footnotetext{\url{https://www.kaggle.com/competitions/web-traffic-time-series-forecasting/data}}
    \item \textbf{Power-cons}, which contains hourly power consumption data from three different distribution networks in Tetouan City, located in northern Morocco.\footnote{\url{https://archive.ics.uci.edu/dataset/849/power+consumption+of+tetouan+city}}  The training data span from 2017-05-01 to 2017-08-07, the validation data extend until 2017-09-30,  and the remaining data up to 2017-12-30 are used for testing. We scale the values by dividing all data points by 1e4.
\end{itemize}
Unlike~\cite{deng2024domain}, we did not experiment with synthetic datasets, as they do not reflect real-world scenarios and our method does not depend on all the assumptions outlined in their paper. For the base models, DLinear and GPT4TS, we did not include external features, as their designs focus on handling only the time series variables.

\section{Implementation Details} We implemented all models using Pytorch~\cite{paszke2019pytorch} 2.4.0 with CUDA 12.1 on NVIDIA TITAN Xp and RTX A6000. 
For all datasets, we use the scaling mechanism following prior studies~\cite{salinas2020deepar,sprangers2023parameter}. We also incorporate a simple data normalization block, reverse instance norm to ensure stability training gives the variance among domains~\cite{kim2021reversible}.
We use Glorot initialization~\cite{glorot2010understanding} to initialize parameters and use the Adam~\cite{kinga2015method} optimizer. We use a batch size of 64 and a dropout rate of 0.3. The learning rate is searched from $\{0.0001, 0.0005, 0.001\}$. The hidden state size is the same for all layers, searched from $\{16, 32, 64\}$. 

\textbf{Structure of the conditional $\beta$-VAE}. For Dlinear and GPT4TS, we use a bidirectional GRU layer followed by two linear readouts to estimate the mean and covariance of the latent distribution~\cite{zhao2023revisiting}. For DeepAR and WaveNet, we replace the GRU with an MLP layer, which has been shown to perform better. We use the same decoder network for all models, consisting of a single linear layer. 

\textbf{Hyperparameter searching}.We balance efficiency and performance as recommended by prior work~\cite{sprangers2023parameter}. Thus, we set the number of hidden layers for DeepAR and the kernel size for WaveNet according to established guidelines~\cite{sprangers2023parameter}. To perform hyperparameter search, we first fix the optimal settings of the other hyperparameters learned from the base model and then search for the best values of hyperparameters. For instance, in our method, we use the dimension of hidden states from the base model, and search $\beta$ from the set $\{1, 5, 10, 15\}$, $\alpha$ from the set $\{0.25, 0.5, 0.75\}$ and a moving average kernel size for time series decomposition from $\{5, 9, 13\}$. This also applies to other baselines.

\textbf{Model selection}. To select the best-trained model for domain generalization tasks, we use the training-domain validation method \cite{gulrajani2020search} and follow the procedures outlined in prior work on time series forecasting~\cite{deng2024domain}. We use 80\% of the domains for training and validation, and the remaining 20\% for testing.
We assess each model’s performance using 5 random seeds and report the average results. We shuffle the training and test domain splits for each seed to evaluate the model's generalization performance under varying conditions.

\section{Evaluation Metrics}
We evaluate the probabilistic forecasting performance using both point and range accuracy metrics~\cite{salinas2020deepar,sprangers2023parameter}. All evaluation scores are computed and averaged across all training/test domains.
\begin{itemize}[leftmargin=*]
    \item 
For point accuracy, we employ the normalized root mean squared error (NRMSE) and symmetric mean absolute percentage error (sMAPE)~\cite{armstrong1985crystal}. The formulas for these calculations are as follows:
\begin{equation}
    \mathit{NRMSE} = \frac{\sqrt{\frac{1}{ h} \sum_{t=T+1}^{T+h}(\mathbf{Y}_{t} - \hat{\mathbf{Y}}_{t})^2 } }{\frac{1}{h}\sum_{t=T+1}^{T+h}|\hat{\mathbf{Y}}_{t}| },
\end{equation}
\begin{equation}
    \mathit{sMAPE} = \frac{1}{h}\sum_{t=T+1}^{T+h} \frac{2|\mathbf{Y}_{t} - \hat{\mathbf{Y}}_{t}|}{|\mathbf{Y}_{t}| + |\hat{\mathbf{Y}}_{t}|}.
\end{equation}
where $\mathbf{Y}, \hat{\mathbf{Y}} \in \mathbb{R}^{N \times h}$ denote the matrices of the actual and predicted forecasting sequences for all samples of size $N$, and $h$ represents the forecasting window size. The subscript $t$ indicates the forecasting time step.

\item For range accuracy, we use the normalized quantile loss function~\cite{salinas2020deepar,deng2024domain}:
\begin{equation}
         Q(q) = \frac{ \sum_{t=T+1}^{T+h} 2 \big |(\mathbf{Y}_{t}-{\hat{\mathbf{Y}}}_{t}^q) \cdot ( \mathbbm{1}_{\mathbf{Y}_{t} \leq {\hat{\mathbf{Y}}}_{t}^q} - q) \big | }{\sum_{t=T+1}^{T+h} |\mathbf{Y}_{t}|},
         \label{eq:qloss}
\end{equation}
where $q$ is the quantile value, $\hat{\mathbf{Y}}^q \in \mathbb{R}^{N\times h}$ is the  prediction matrix for  quantile $q$. $\mathbbm{1}$ is the indicator function. We use results when $q = 0.5$, referred as Q(0.5), and the mean quantile performance across nine quantiles $q = \{0.1, 0.2, \ldots, 0.9\}$, referred as Q(mean).

\end{itemize}

\section{Additional Model Analyses}
\subsection{Effect of Hyperparameter $\beta$} 
\begin{figure}[ht]
  \centering 
  \begin{subfigure}{.4\textwidth}
    \centering
\includegraphics[width=1.0\linewidth]{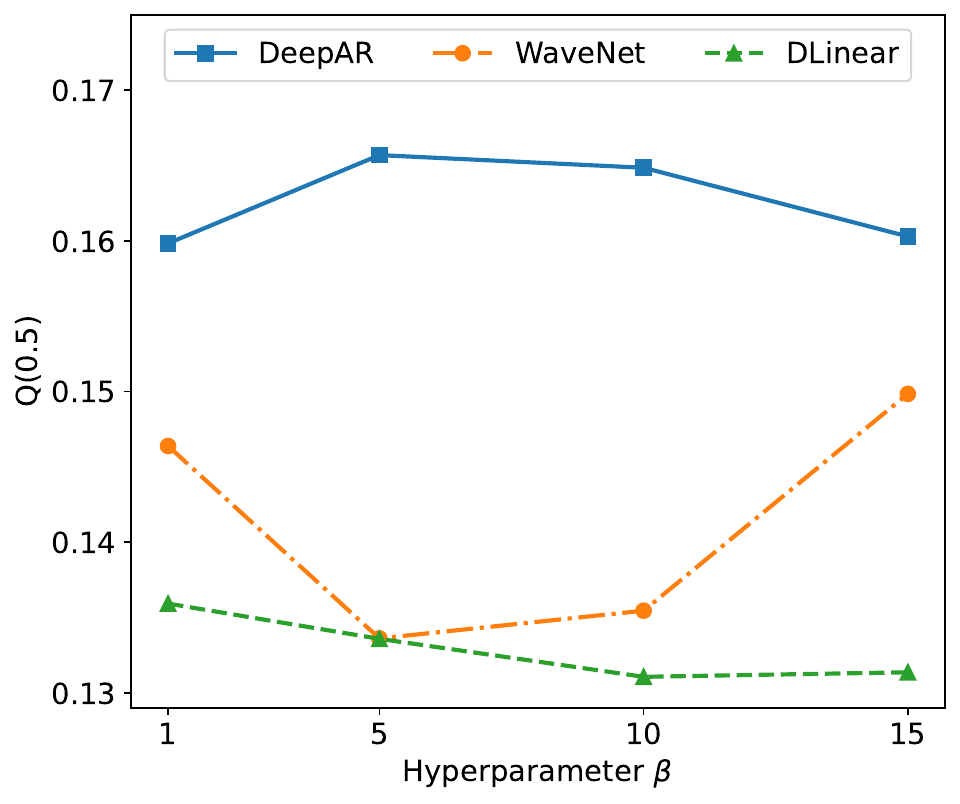}
  \end{subfigure} 
 \hspace{10pt}
    \begin{subfigure}{.4\textwidth}
    \centering
\includegraphics[width=1\linewidth]{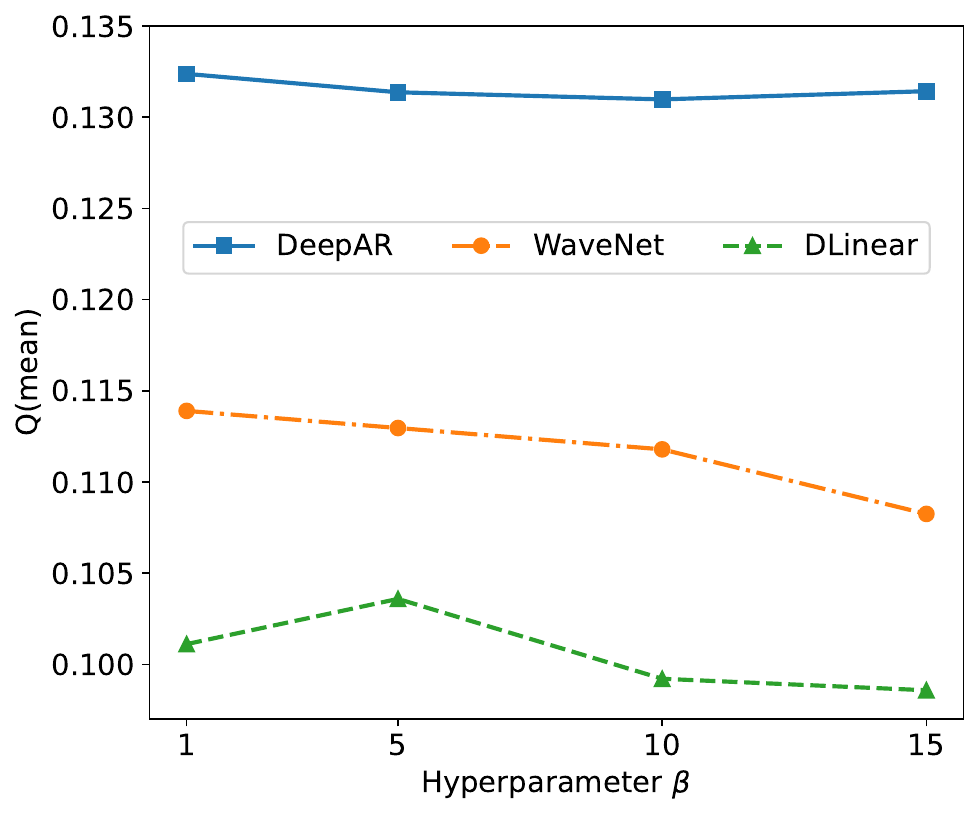}
  \end{subfigure} 
 \caption{Forecasting results of Q(0.5) and Q(mean) on the Web-traffic dataset with varying values of $\beta$.} 
  \label{fig:beta-analysis}
 \end{figure}

We use $\beta$ to control the learning of latent vectors, with the goal that each dimension of the latent vector captures independent factors~\cite{higgins2017beta,burgess2018understanding}. To evaluate the effect of $\beta$, we analyze the forecasting performance across different base models using our proposed method. Experiments are conducted on the Web-traffic dataset, and the results of the range accuracy metrics are shown in Figure~\ref{fig:beta-analysis}. 
We observe that larger $\beta$ values generally improve forecasting performance, though the impact varies slightly across models. However, when $\beta$ becomes too large (e.g., $\beta = 15$), the performance of WaveNet drops (e.g., Q(0.5) increases indicating the central trend of the data is not accurately estimated), which may be due to over-regularization, causing the latent representations to lose essential information needed for accurate forecasting. It would be beneficial to adjust the value for different real-world datasets, given their versatility.

\subsection{Additional Latent Space Analysis} 
\begin{figure}[ht]
  \centering 
  \begin{subfigure}{.36\textwidth}
    \centering
\includegraphics[width=1.0\linewidth]{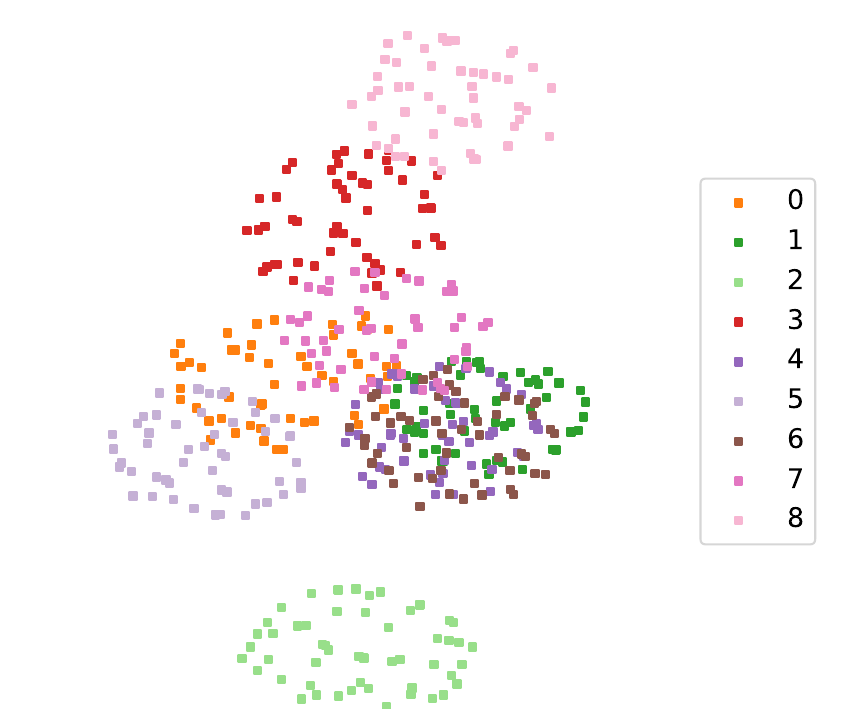}
 \hspace{10pt}

  \end{subfigure} 
    \begin{subfigure}{.36\textwidth}
    \centering
\includegraphics[width=1\linewidth]{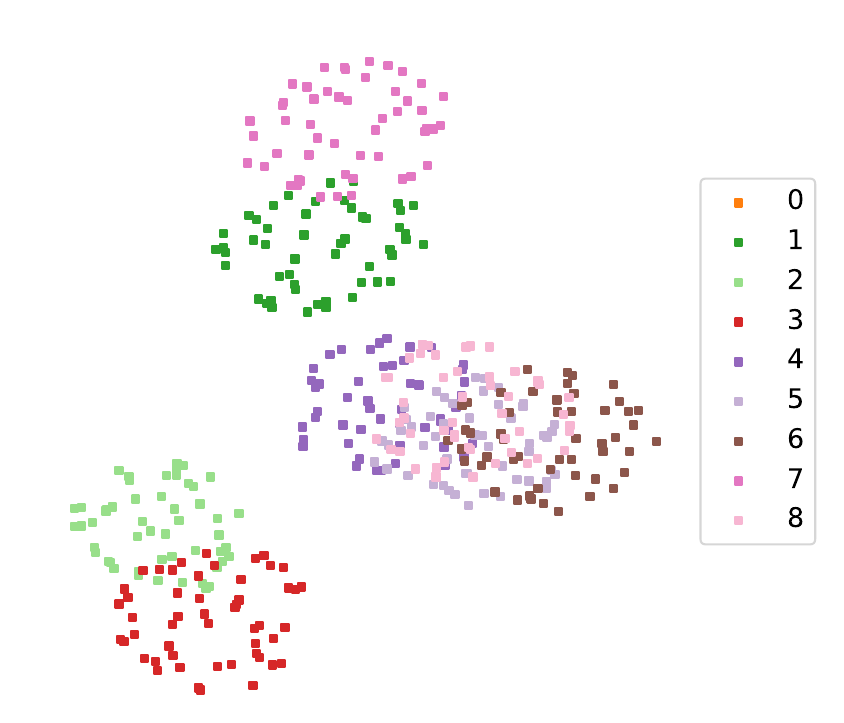}
  \end{subfigure} 
  \begin{subfigure}{.36\textwidth}
    \centering
\includegraphics[width=1.0\linewidth]{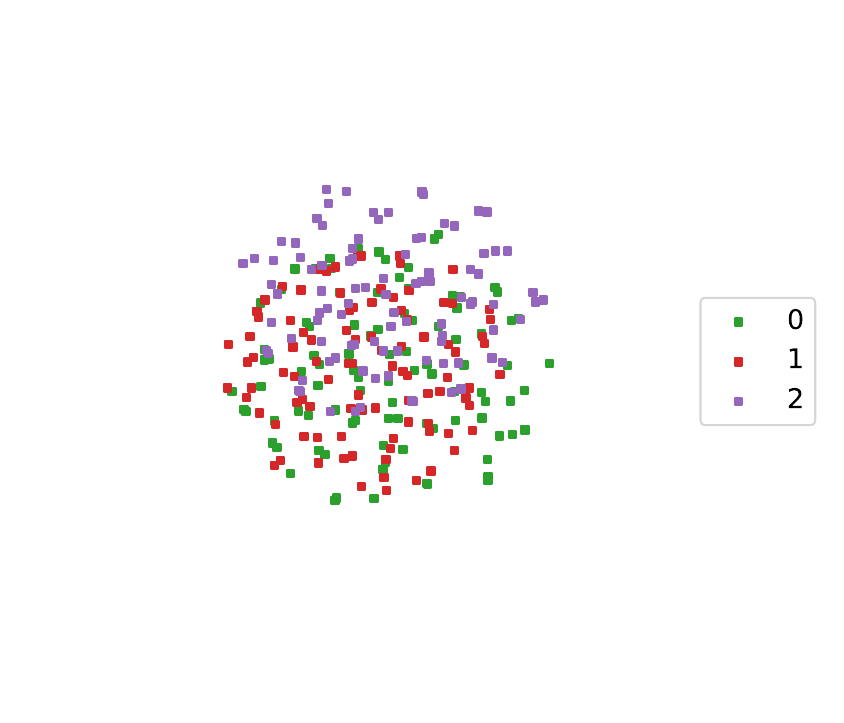}
\vspace{-10pt}
\caption{Domain-shared component}
  \end{subfigure} 
 \hspace{10pt}
    \begin{subfigure}{.36\textwidth}
    \centering
\includegraphics[width=1\linewidth]{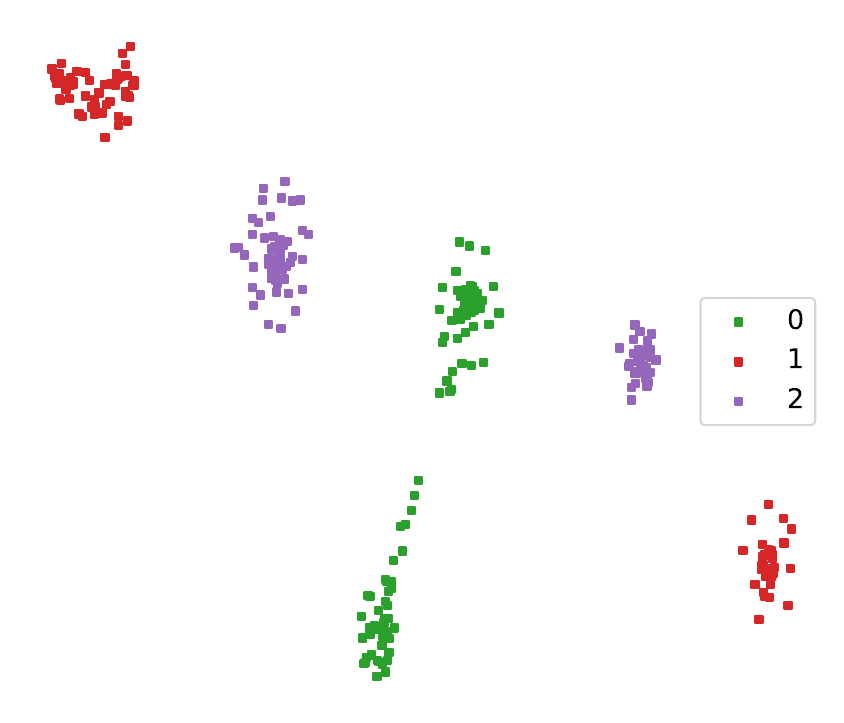}
\vspace{-10pt}
\caption{Domain-specific component}
  \end{subfigure} 
 \caption{Latent space t-SNE visualization of the domain-shared and domain-specific components on the test domains of the Favorita-store (top) and Stock-volume (bottom) datasets (from top to bottom).} 
  \label{fig:more-latent-analysis}
 \end{figure}

We present more examples of latent spaces learned by our model based on DeepAR, as shown in Figure~\ref{fig:more-latent-analysis}.

\subsection{Additional Showcases} 
\begin{figure*}[h]
  \centering 
  \begin{subfigure}{.19\textwidth}
    \centering
\includegraphics[width=1.0\linewidth]{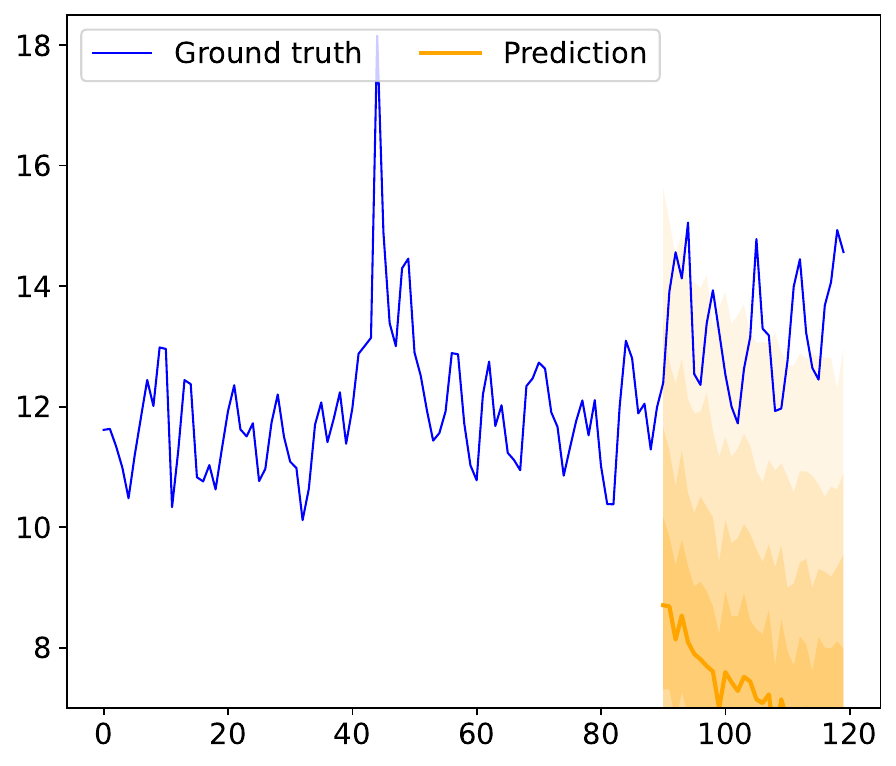} 
  \end{subfigure}
    \begin{subfigure}{.19\textwidth}
    \centering
\includegraphics[width=1\linewidth]{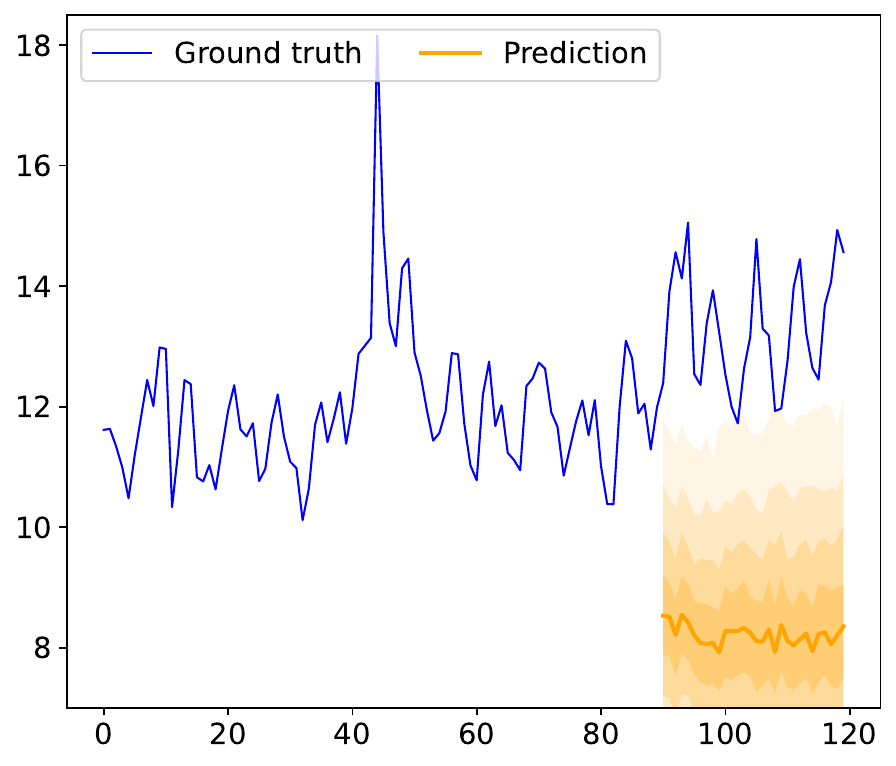} 
  \end{subfigure} 
    \begin{subfigure}{.19\textwidth}
    \centering
    \includegraphics[width=1.0\linewidth]{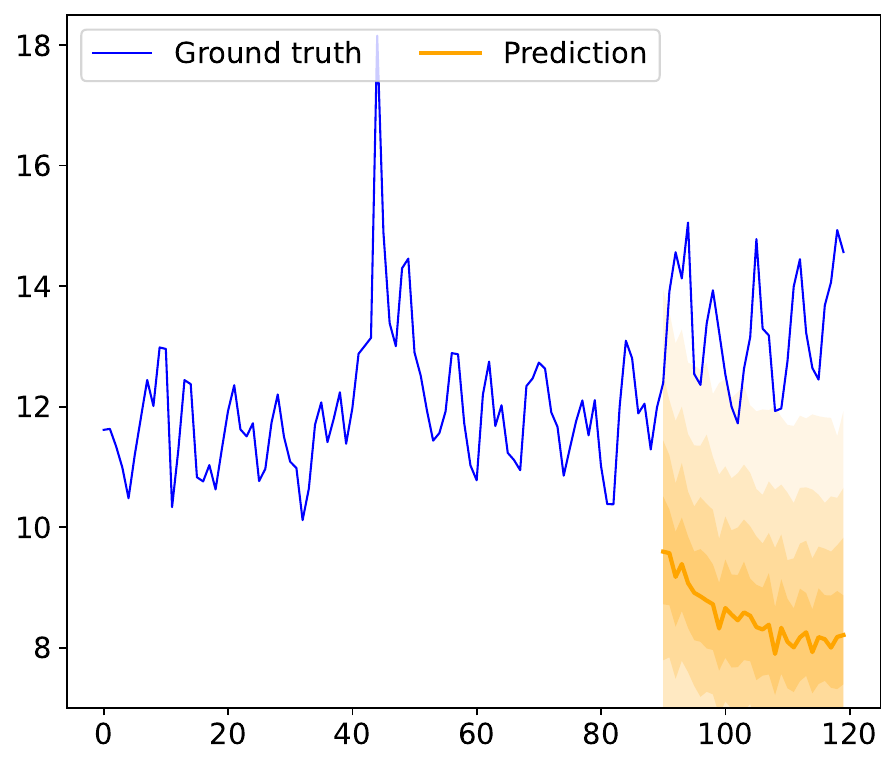} 
  \end{subfigure} 
  \begin{subfigure}{.19\textwidth}
    \centering
\includegraphics[width=1.0\linewidth]{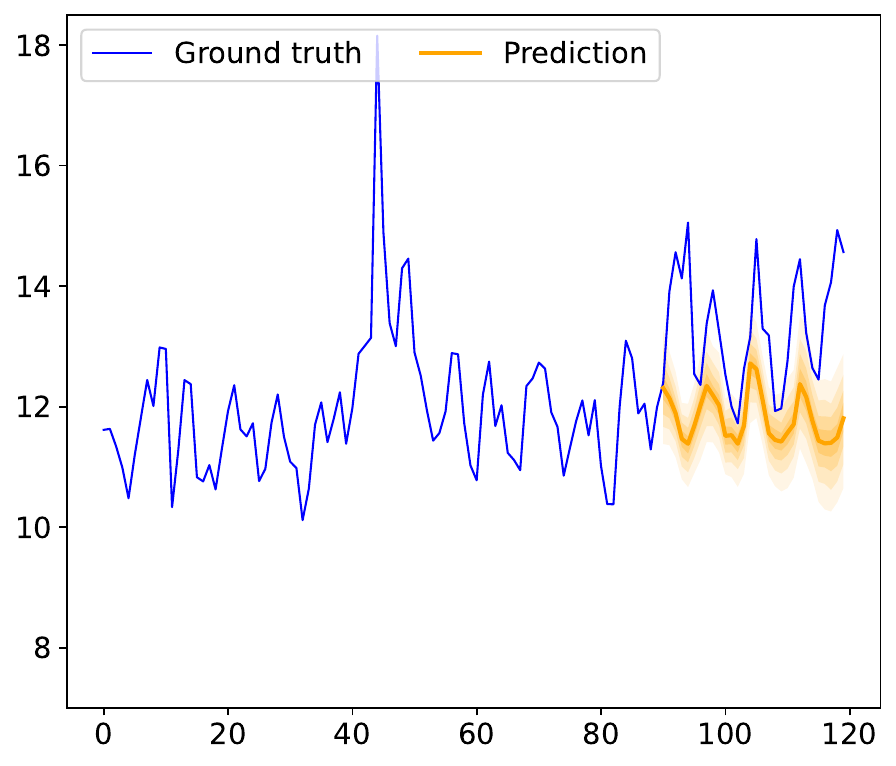} 
  \end{subfigure} 
    \begin{subfigure}{.19\textwidth}
    \centering
\includegraphics[width=1.0\linewidth]{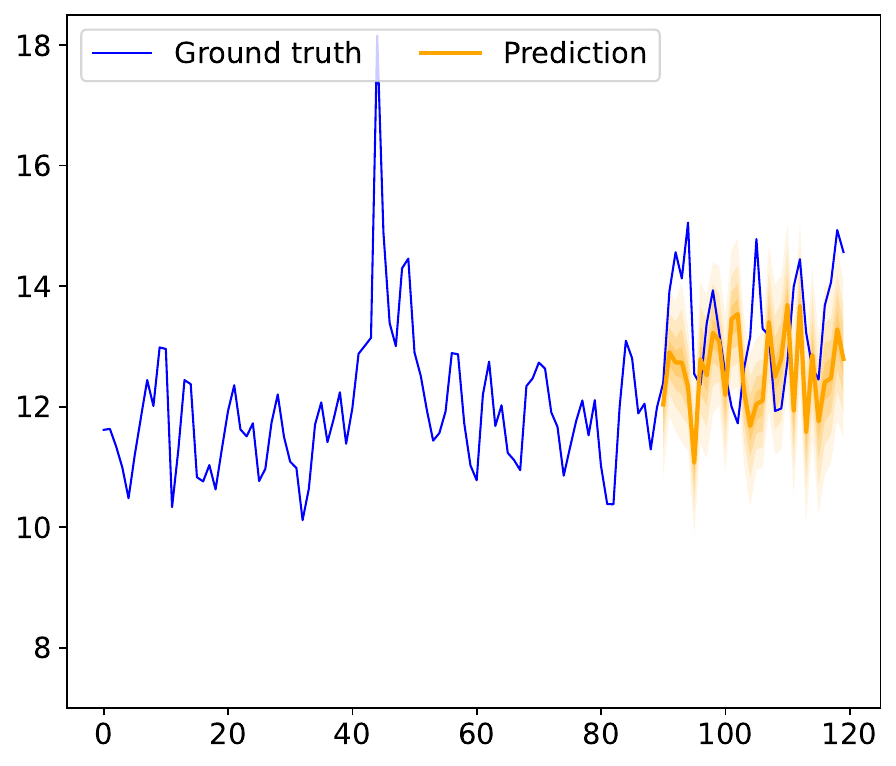}
  \end{subfigure} 

    \begin{subfigure}{.19\textwidth}
    \centering
\includegraphics[width=1.0\linewidth]{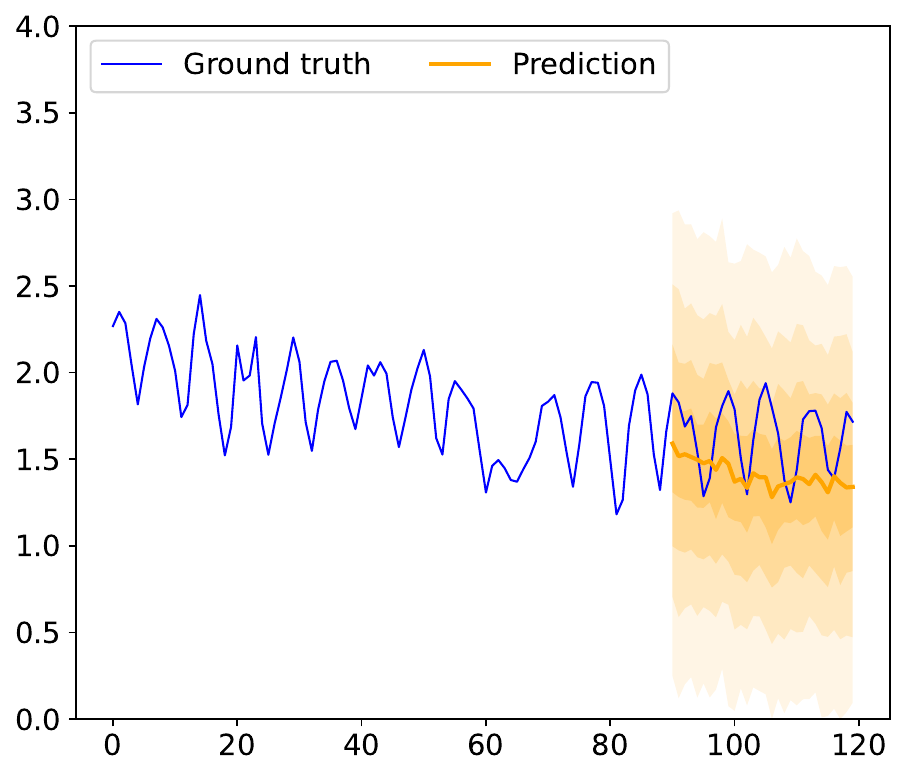}\caption{WaveNet}
  \end{subfigure}
    \begin{subfigure}{.19\textwidth}
    \centering
\includegraphics[width=1\linewidth]{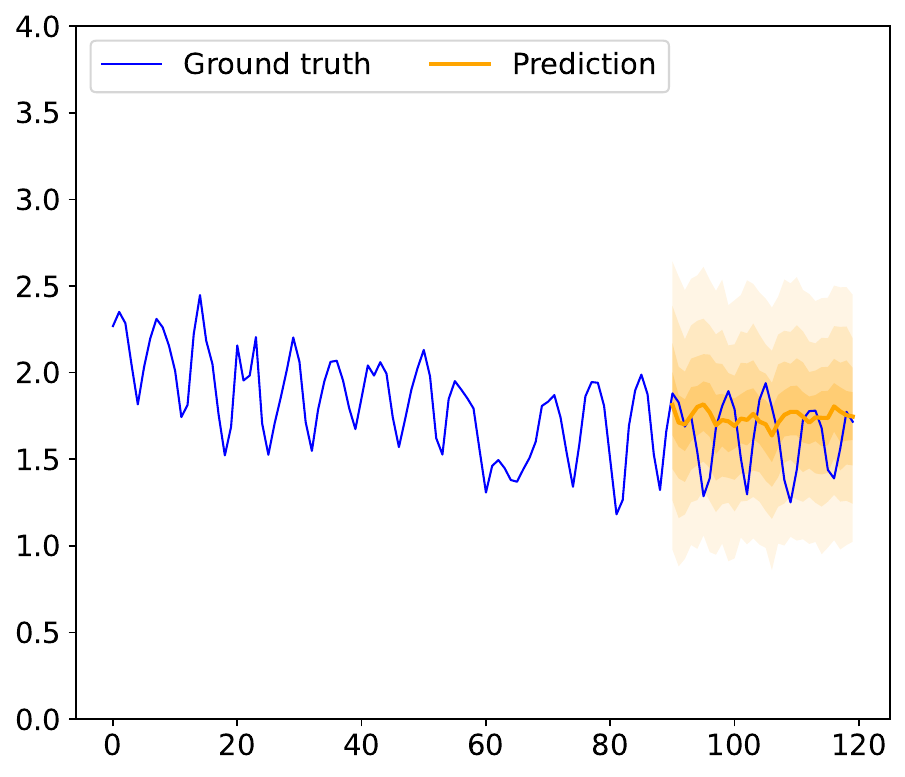}\caption{WaveNet+IDGM}
  \end{subfigure} 
    \begin{subfigure}{.19\textwidth}
    \centering
    \includegraphics[width=1.0\linewidth]{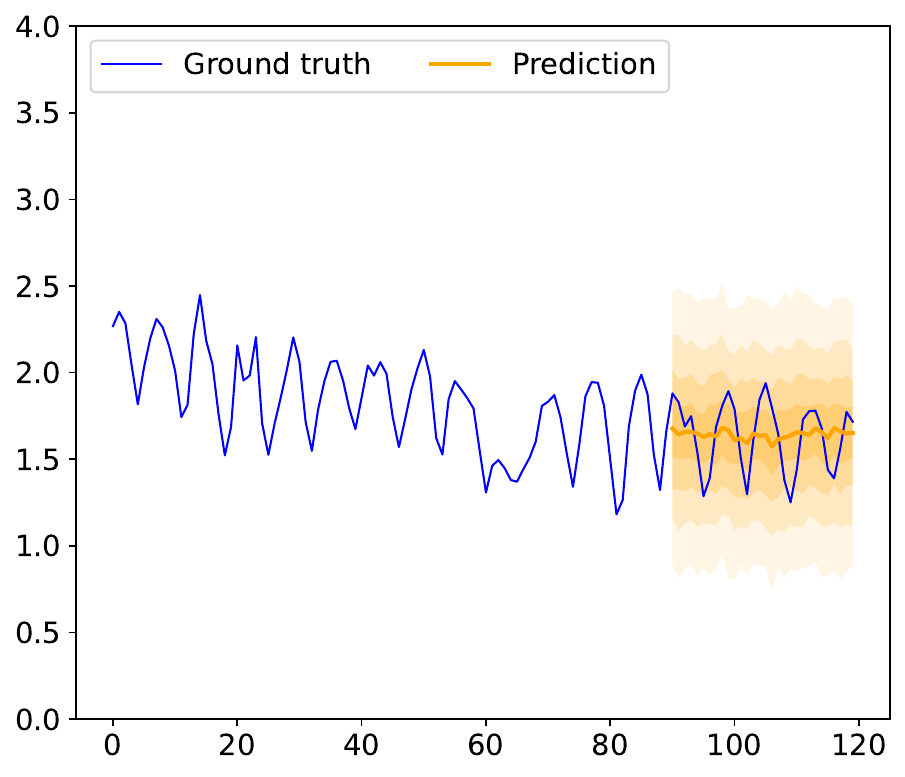}\caption{WaveNet+Cedar} 
  \end{subfigure} 
  \begin{subfigure}{.19\textwidth}
    \centering
\includegraphics[width=1.0\linewidth]{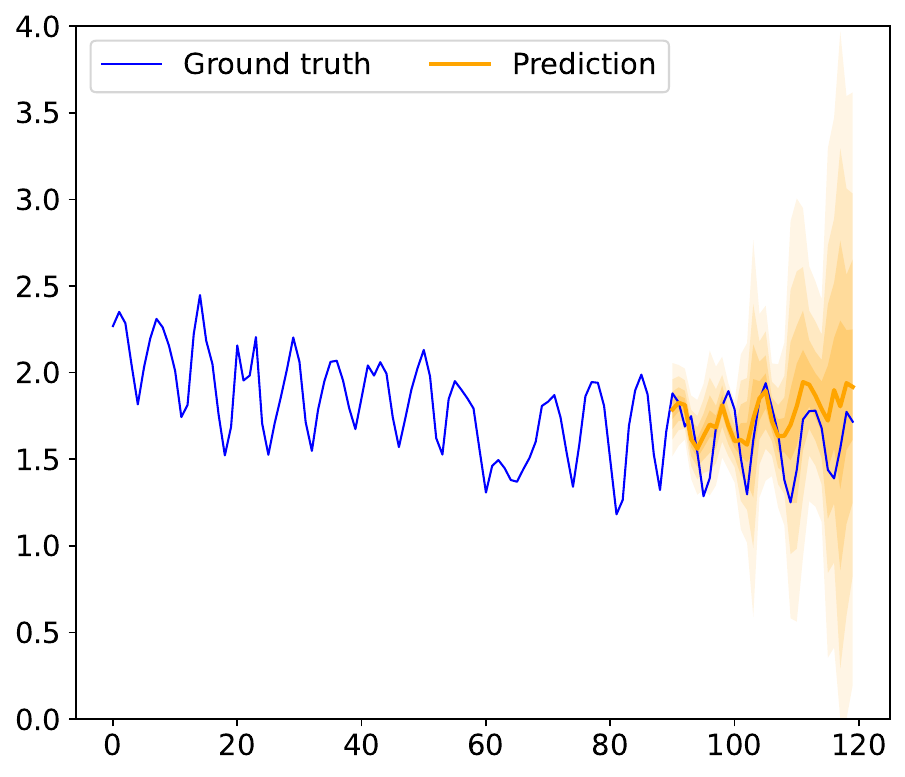}\caption{WaveNet+\ours} 
  \end{subfigure} 
    \begin{subfigure}{.19\textwidth}
    \centering
\includegraphics[width=1.0\linewidth]{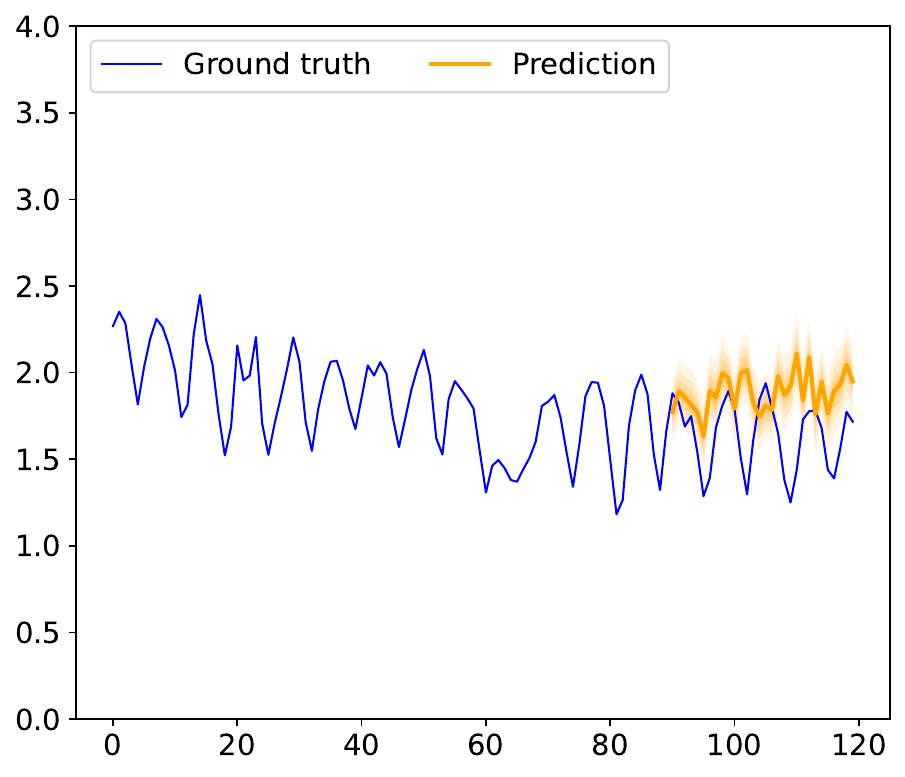}\caption{DLinear}
  \end{subfigure} 

 \caption{Forecasting results for test domain samples from the Web-traffic dataset, using DLinear and WaveNet with various generalization methods. Each row represents a different sequence example.} 
  \label{fig:more-showcases}
 \end{figure*}

\begin{figure*}[ht]
  \centering 
  \begin{subfigure}{.19\textwidth}
    \centering
\includegraphics[width=1.0\linewidth]{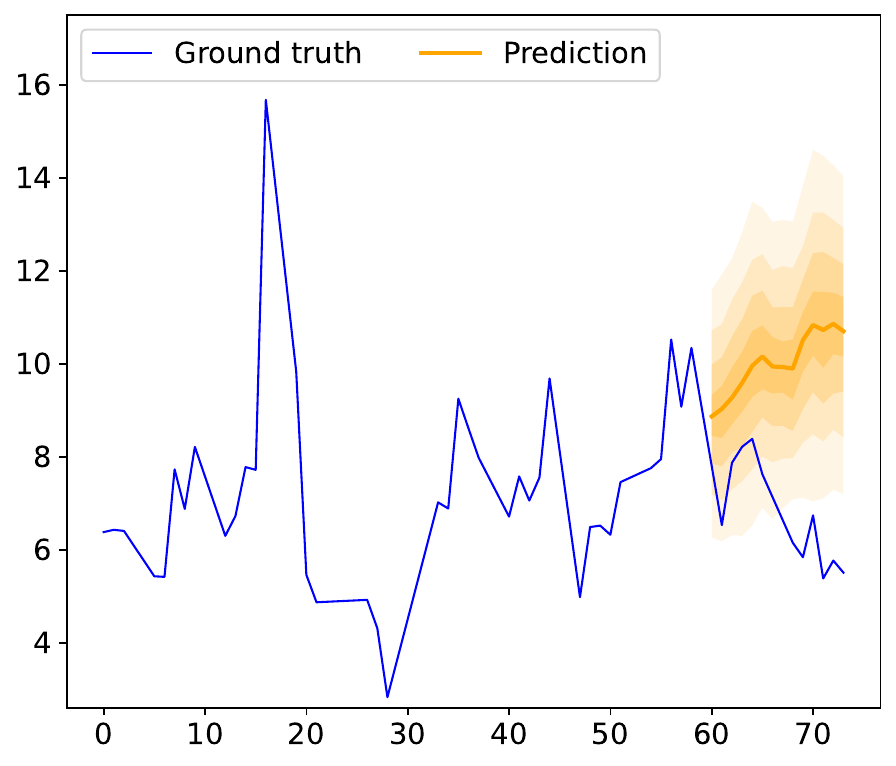}\caption{DeepAR}
  \end{subfigure}
    \begin{subfigure}{.19\textwidth}
    \centering
\includegraphics[width=1\linewidth]{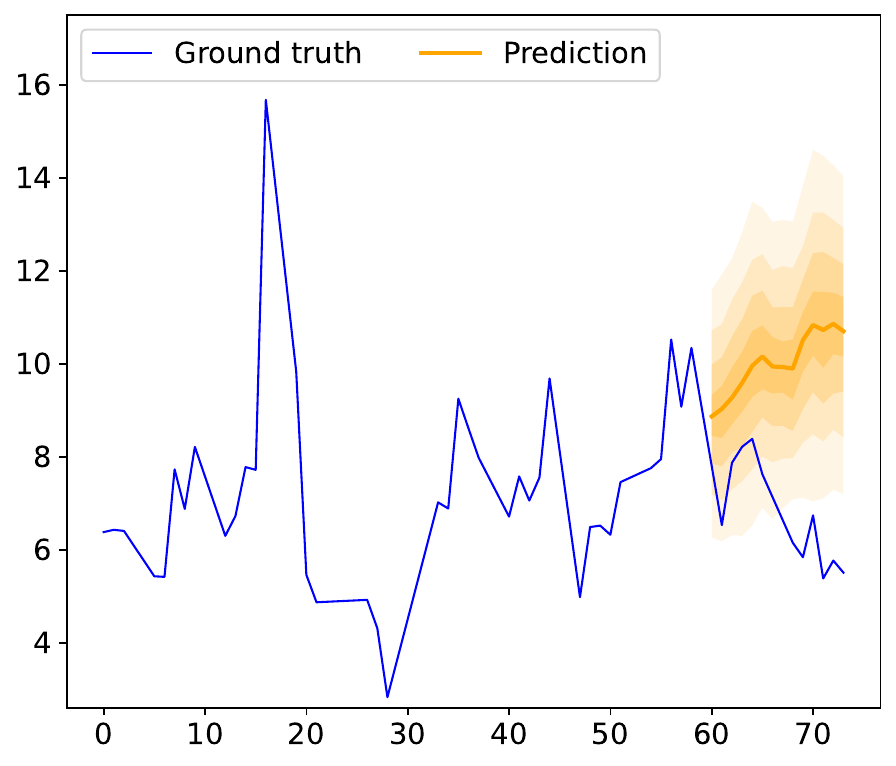}\caption{DeepAR+IDGM}
  \end{subfigure} 
    \begin{subfigure}{.19\textwidth}
    \centering
    \includegraphics[width=1.0\linewidth]{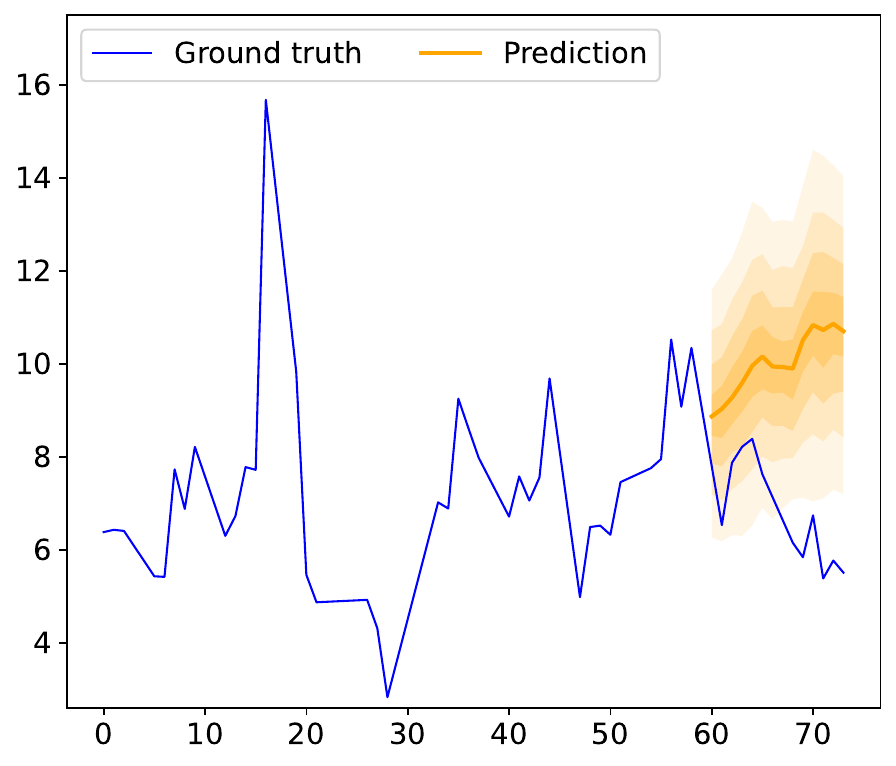}\caption{DeepAR+Cedar}
  \end{subfigure} 
  \begin{subfigure}{.19\textwidth}
    \centering
\includegraphics[width=1.0\linewidth]{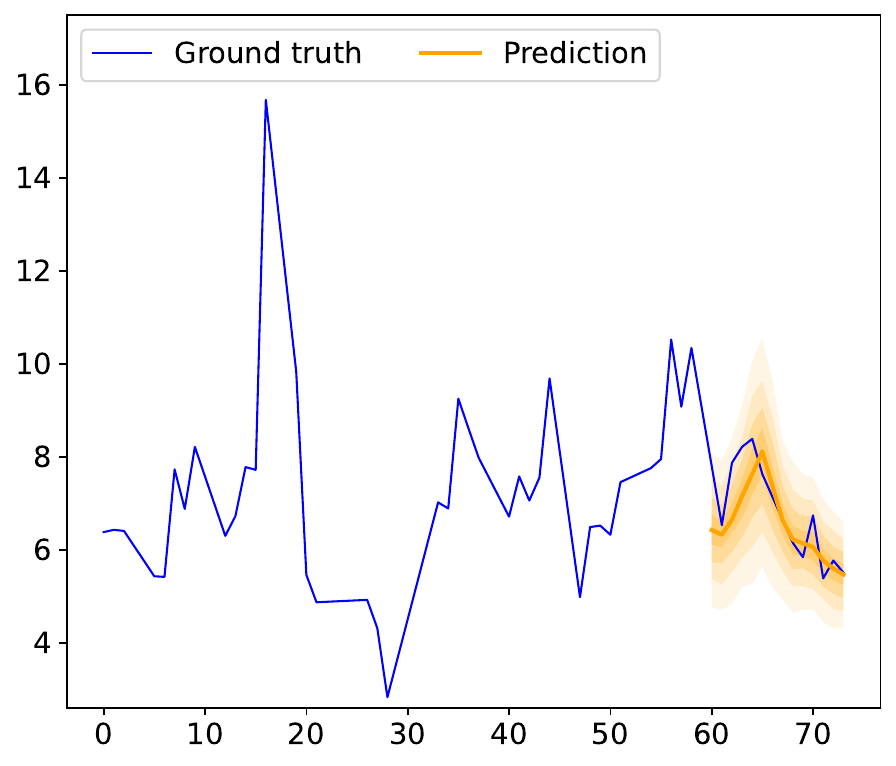}\caption{DeepAR+\ours}
\end{subfigure}
 \caption{Forecasting results for a test domain sample from the Stock-volume dataset, using DeepAR with various generalization methods.} 
  \label{fig:more-showcases-stock}
 \end{figure*}

We present additional visualizations of predicted sequences in Figures~\ref{fig:more-showcases}
and \ref{fig:more-showcases-stock}. It is evident that \ours{} demonstrates the strongest ability to capture the movement of time series data from relevant test domains.

\subsection{Additional LLM-based Models}
There are some LLM-based approaches, such as TimeLLM~\cite{jin2024timellm}, Chronos~\cite{ansari2024chronos}, and Uni2TS~\cite{woo2024unified}, have been developed for time series forecasting. 
TimeLLM adapts frozen large language models for time series forecasting using a reprogramming approach. Chronos and Uni2TS are large transformer-based models trained on a large volume of time series data.
We experimented with these methods using their open-source code and presented the results in Table~\ref{tab:res-llm}. We use pretrained Chronos and Uni2TS models.
We found that these methods struggle to achieve state-of-the-art performance on our problem, likely due to design misalignment and susceptibility to overfitting on small datasets. They also come with high computational costs. Therefore, we did not include them as baselines in this paper.

\setlength{\tabcolsep}{3.1pt} 
\begin{table*}[ht]
\centering
\caption{Forecasting results of additional LLM-based methods on three real-world datasets.}
\label{tab:real-res}
\rotatebox{0}{\begin{tabular}{l cc cc cc}
\toprule
       & 
       \multicolumn{2}{c}{\textbf{Favorita-cat}}  & \multicolumn{2}{c}{\textbf{Favorita-store}}    & \multicolumn{2}{c}{\textbf{Stock-volume}} \\
        & Q(0.5)           & sMAPE       & Q(0.5)           & sMAPE   & Q(0.5)           & sMAPE        \\
        \midrule
Chronos-t5-mini & 0.1180 \tiny{.0463} & 0.2725 \tiny{.1306} & 0.0265 \tiny{.0018} &  0.0264 \tiny{.0018} & 0.1966 \tiny{.0429} & 0.1974 \tiny{.0360} \\
Uni2TS-base  & 0.1481 \tiny{.0134} & 0.3682 \tiny{.0373} & 0.0331 \tiny{.0004} & 0.0651 \tiny{.0005} & 0.2248 \tiny{.0181} & 0.2183 \tiny{.0068} \\
TimeLLM & - & 0.1910 \tiny{.0683} & - & 0.0249 \tiny{.0046} & - & 0.1990 \tiny{.0259} \\
\bottomrule
\end{tabular}}\label{tab:res-llm}
\end{table*}

\end{document}